\documentclass{article} % For LaTeX2e
\usepackage{iclr2027_conference,times}

\usepackage{amsmath,amsfonts,bm}

\def\eqref#1{equation~\ref{#1}}
\def\1{\bm{1}}

\DeclareMathAlphabet{\mathsfit}{\encodingdefault}{\sfdefault}{m}{sl}
\SetMathAlphabet{\mathsfit}{bold}{\encodingdefault}{\sfdefault}{bx}{n}

\usepackage{hyperref}
\usepackage{url}
\usepackage{graphicx}
\usepackage{booktabs,multirow}
\usepackage{xspace}
\usepackage{bbm}
\usepackage{enumitem}
\usepackage{wrapfig}
\usepackage{subcaption}
\usepackage{amssymb}

\title{Sleep2Word: Peripheral-to-EEG Translation via Duration-Aware Wordization}

\author{Keondo Park$^a$, Arthur Delhaye$^b$, Younghoon Na$^{c,d}$, Hyun-Woo Shin$^{c,d,e}$, and Hyung-Sin Kim$^a$$\dagger$\\
\textnormal{\small $^a$Graduate School of Data Science, Seoul National University, South Korea}\\
\textnormal{\small $^b$Computer Science, Université du Québec à Rimouski, Canada}\\
\textnormal{\small $^c$Department of Biomedical Sciences, Seoul National University College of Medicine, Republic of Korea}\\
\textnormal{\small $^d$Obstructive Upper Airway Research (OUaR) Laboratory, Department of Pharmacology}\\
\textnormal{\small \quad Seoul National University College of Medicine, Republic of Korea}\\
\textnormal{\small $^e$Department of Otorhinolaryngology-Head and Neck Surgery, Seoul National University Hospital, Republic of Korea}}

\newcommand{\edit}[1]{\textcolor{black}{#1}}
\newcommand{\proposal}{\textit{Sleep2Word}\xspace}

\iclrfinalcopy % Uncomment for camera-ready version, but NOT for submission.
\begin{document}

\maketitle

\begin{abstract}
Estimating full-night electroencephalography (EEG) spectrograms from peripheral measurements could extend EEG-informed sleep analysis to recordings acquired without EEG. This task requires compact representations of signals with different temporal structures. 
We introduce \proposal, a framework for translating respiration, electrocardiography (ECG), or photoplethysmography (PPG) into EEG spectrograms through duration-aware physiological words. Building on context-dependent chunking, \proposal groups physiological embeddings into variable-duration chunks and jointly quantizes their content and duration into modality-specific wordbooks. 
A block-autoregressive translator then predicts target EEG word identities and occurrence-specific duration refinements within fixed-duration blocks.
This design accommodates independent source and target wordbooks, limits temporal drift, and shortens the autoregressive decoding path.
% To place generated EEG words on the recording timeline, a block-autoregressive decoder predicts word identities and duration refinements within fixed-duration intervals. 
% The model combines full-night source context with sequential target prediction within intervals and parallel generation across intervals, allowing source and target modalities to retain different segmentations.
%
Trained on over 20K PSG cases across 9 datasets, \proposal consistently outperforms existing spectrogram generation baselines in EEG spectrogram reconstruction, sleep staging, arousal detection, and survival prediction.
Compressing full night sequences by up to 69.7\%, the learned words yield a favorable accuracy–compute trade-off. 
Sleep event analysis further shows that the learned words capture physiologically plausible associations across modalities, suggesting their potential for interpretable cross-modal analysis.

\end{abstract}

\section{Introduction}

Electroencephalography (EEG) provides a detailed view of sleep-related brain activity, but acquiring it routinely outside a sleep laboratory remains challenging. Cardiac and respiratory measurements are more accessible and carry partial information about brain state through shared sleep physiology~\citep{jurysta2003study, schreiner2023respiration}. 
We study \textbf{peripheral-to-EEG translation}: estimating a full-night EEG spectrogram from an individual peripheral source, respiration, electrocardiography (ECG), or photoplethysmography (PPG), without EEG at inference. The target is spectral structure that can be predicted from the observed peripheral signal, not exact recovery of an unobserved EEG waveform. %An EEG spectrogram also provides a common intermediate output for studying multiple downstream tasks, rather than restricting the cross-modal mapping to a single task-specific label. 
The estimated spectrogram serves as a common intermediate output for multiple downstream analyses, rather than limiting the peripheral-to-brain mapping to a single task-specific label.

Physiology as Language~\citep{zha2026physiology} established the feasibility of respiration-to-sleep-EEG synthesis using continuous respiratory conditioning, a fixed grid of discrete EEG tokens, and masked generative prediction. 
We build on this task by investigating how both source and target signals should be represented. Fixed-duration tokenization provides a convenient temporal grid but uses the same segmentation granularity throughout a recording. Short intervals produce long full-night sequences, whereas longer intervals provide a coarser representation. Different physiological modalities also exhibit distinct temporal structures, motivating segmentations that can v\textit{ary both within and across signals}. 
We therefore ask whether learned, variable-duration segments can reduce the cost of full-night translation while retaining the peripheral information needed to estimate EEG. This motivates \textbf{adaptive representations on both sides of the translation}, rather than prescribing one segment duration throughout each recording.

% However, short patches produce long sequences for full-night recordings, whereas longer patches represent the signal through fewer, wider intervals. 
% This motivates an alternative: learning variable-duration segments whose lengths depend on the local signal, rather than selecting one patch duration for the entire recording. 
% Such a representation seeks to shorten the sequence while retaining the signal content needed for translation. 

Adaptive segmentation makes duration part of the translation problem. The fifth respiratory token and the fifth EEG token need not describe the same recording interval, even when the underlying signals are synchronized. Although source timestamps remain known, the decoder must predict both the content and duration of the target segments. In a continuous autoregressive chain, accumulated duration errors can shift later predictions along the recording. We therefore seek modality-specific segmentations whose translation remains referenced to a common physical timeline, without requiring corresponding physiological patterns to occur simultaneously.

\begin{figure*}[t]
    \vspace{-2ex}
    \centering
    \includegraphics[width=0.995\textwidth,bb=0 0 548 233]{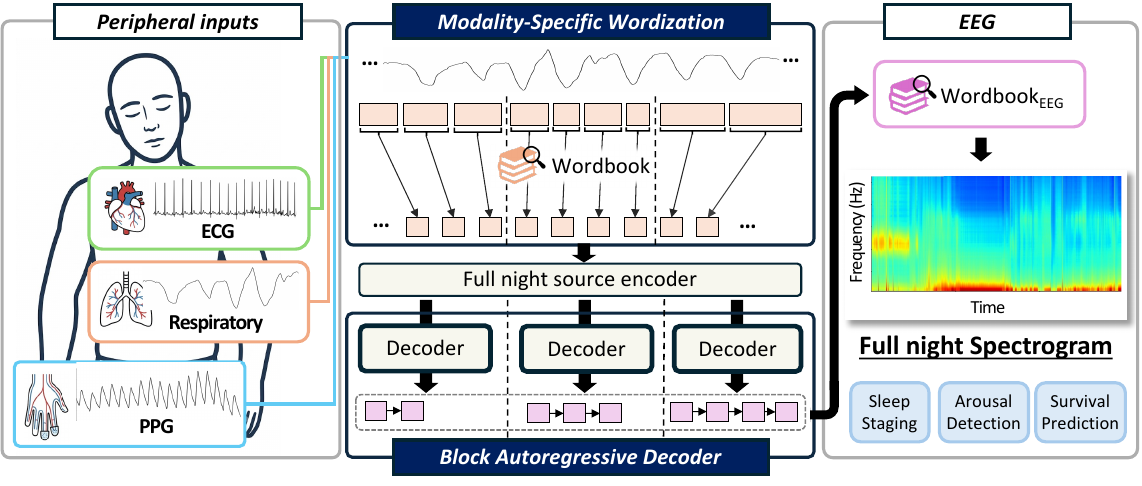}
    \vspace{-2ex}
    \caption{\textbf{Modality-specific physiological words on a shared recording timeline}. \proposal represents peripheral and EEG signals using independently learned, variable-duration words. 
    %Common recording intervals organize translation without requiring identical word boundaries across modalities. 
    EEG words are generated autoregressively within block, regularizing duration drift and enabling parallel generation of different blocks.}
    \label{fig:overview}
    \vspace{-5ex}
\end{figure*}

We introduce \proposal (Figure~\ref{fig:overview}), guided by a principle: let each modality learn its own segments, but organize translation around common recording intervals. 
\proposal adapts context-dependent chunking~\citep{hwang2026dynamic} to physiological embeddings from SleepMaMi~\citep{park2026sleepmami}, grouping adjacent embeddings into variable-duration chunks under reconstruction supervision. 
\textbf{Duration-aware wordization} then quantizes each chunk's pooled representation together with its duration into a modality-specific wordbook. 
A physiological word is a learned discrete prototype of chunk content and temporal extent. Chunking shortens the sequence, while wordization expresses translation as prediction over reusable segment codes. Because occurrences can differ from a word's typical duration, the decoder predicts an occurrence-specific duration adjustment.

%let each modality learn its own segments, but organize translation around common recording intervals. 
%\proposal first adapts context-dependent chunking ~\citep{hwang2026dynamic} to physiological embeddings from SleepMaMi ~\citep{park2026sleepmami}, grouping adjacent embeddings into variable-duration segments under reconstruction supervision. 
%It then performs \textbf{duration-aware wordization}, discretizing each segment’s pooled embedding together with its duration into a modality-specific wordbook. 
%Content and duration therefore jointly determine word identity. 
%Chunking shortens the source and target sequences, while wordization provides discrete units for translating between them without requiring matching token counts or segment lengths. 

To translate these words, a full-night source encoder first captures context across the recording. A \textbf{block-autoregressive decoder} then generates EEG words and their duration adjustments within fixed-duration \emph{blocks}, with sequential prediction inside each block and parallel generation across blocks. 
% During chunking, a boundary regularizer encourages segment boundaries at these same temporal intervals. 
The blocks organize generation in time, while the words retain adaptive segmentation within them. Compared with an uninterrupted full-night autoregressive chain, this design shortens the sequential decoding path and is intended to limit cumulative timing errors. Tokenization and generation are therefore coordinated through a common temporal partition.

%To translate these words, \proposal uses \textbf{block-autoregressive generation}. 
%Each decoding block corresponds to a fixed-duration recording interval. 
%The decoder predicts EEG words autoregressively within a block, modeling dependencies among its target words, while multiple blocks can be generated in parallel. 
%Crucially, the segmentation stage is coordinated with this decoding structure: a boundary regularizer encourages word boundaries at the same recording intervals used to define the blocks. 
%The blocks thus organize generation in time, while learned word boundaries can vary within them. 
%This design is intended to limit cumulative temporal drift without extending a single autoregressive decoding chain across the entire night. 
%It combines adaptive representation and temporal organization rather than treating them as independent stages. 

Our contributions are threefold:
\begin{itemize}[leftmargin=1.0em, itemsep=0pt, topsep=0pt, parsep=0pt, partopsep=0pt]
    \item \textbf{Duration-aware physiological wordization}. 
    We adapt context-dependent chunking and jointly quantize chunk content and duration to construct modality-specific wordbooks for peripheral-to-EEG translation. The representation supports variable-duration source and target sequences without imposing matching segmentation.
    \item \textbf{Temporally anchored  word-to-word translation}. 
    We adapt block-autoregressive decoding to physical recording intervals, complemented by occurrence-specific duration prediction. This confines temporal drift of variable-length words within individual blocks and keeps the source and target aligned on the recording timeline.
    % % 수정 each generated EEG word -> generated EEG word
    % We coordinate segment boundaries with block-autoregressive decoding and refine the durations of generated EEG words. This combines local target dependencies and parallel block generation with a common recording-time structure.
    % Block autoregressive decoding and occurrence-specific duration prediction complements using variable length word, by limiting the temporal drift within individual block, aligning source and target recording timeline. 
    \item \textbf{Evaluation across peripheral sources and downstream tasks}. 
    We evaluate respiration-, ECG-, and PPG-to-EEG translation across SHHS1 and MESA as well as a held-out dataset CFS. \proposal outperforms existing baselines in spectrogram reconstruction, downstream tasks including sleep staging, arousal detection and survival prediction, while demonstrating superior computational efficiency. %\proposal improves [PRIMARY METRIC] by [VALUE] over [BASELINE] while reducing sequence length by 80.2\% relative to [REFERENCE TOKENIZATION].
\end{itemize}

\section{Related Work}
\subsection{Cross-System Physiological Coupling and Cross-Modal Modeling}
Accumulating evidence indicates that central, autonomic, and respiratory processes are coupled across multiple timescales. 
% During wakefulness, nasal respiration entrains oscillatory activity in limbic brain regions \citep{zelano2016nasal}. 
During sleep, respiration has been shown to modulate the timing of non-rapid-eye-movement (NREM) slow oscillations and sleep spindles, such that these events preferentially occur at specific phases of the respiratory cycle \citep{schreiner2023respiration}.
In addition, cortical slow oscillations are temporally coupled with transient heart-rate bursts during sleep, demonstrating an interaction between central and autonomic activity \citep{naji2019coupling}. 
Together, these findings suggest that peripheral physiological signals retain information about brain states through shared physiological regulation.

Physiological machine learning exploits cross-system dependencies for representation learning, direct prediction, and signal generation. Sleep foundation models learn reusable representations from multimodal recordings~\citep{park2026sleepmami,thapa2026multimodal}, while xMAE uses masked cross-modal reconstruction as supervision for biosignal representation learning~\citep{zhou2026physiologyaware}. Direct prediction methods such as wav2sleep~\citep{carter2025wav} instead map physiological inputs to sleep-stage labels. 
Cross-modal signal generation methods includes PPG-to-ECG~\citep{ding2025ai}, PPG-to-respiration~\citep{shuzan2024ppg2respnet}, and respiration-to-EEG~\citep{zha2026physiology} synthesis.
%Such cross-system dependencies have increasingly been exploited in physiological machine learning. 
%Recent sleep foundation models~\citep{park2026sleepmami,thapa2026multimodal} leverage paired multimodal recordings to learn modality-shared representations, for example through contrastive objectives or cross-modal reconstruction. 
%Beyond representation learning, several studies have investigated direct cross-modal signal translation.
%For example, PPG-to-ECG models~\citep{ding2025ai,zhou2026physiologyaware} reconstruct ECG-like waveforms from optical pulse signals by exploiting their shared cardiovascular origin. 
%Similarly, PPG2RespNet~\citep{shuzan2024ppg2respnet} synthesizes respiratory waveforms directly from raw PPG using a nested U-Net-inspired architecture.
Most closely related, Physiology as Language~\citep{zha2026physiology} combines continuous respiratory conditioning with a fixed temporal grid of discrete EEG codes. 
Instead, \proposal represents both source and target through \textit{modality-specific words}. It investigates whether this word-based compact representation can support accurate cross-modal estimation. %when word identity and generated duration are modeled together and decoding is organized around common recording intervals.

\subsection{Adaptive Chunking and Discrete Signal Representations}
Adaptive sequence models reduce fine-grained processing by forming input-dependent segments. SpaceByte~\citep{slagle2024spacebyte} concentrates higher-capacity processing around delimiter-like boundaries, while BLT~\citep{pagnoni2025byte} forms variable-length byte patches using predictive entropy. H-Net~\citep{hwang2026dynamic}  learns context-dependent boundaries through a trainable routing module. 
\proposal builds on this learned chunking mechanism and adapts it using physiological reconstruction and synchronization-oriented boundary regularization.

Discrete signal vocabularies have also been studied beyond language. TFM-Tokenizer~\citep{pradeepkumar2026tokenizing} learns discrete time–frequency motifs from single-channel EEG. 
In speech, VARSTok~\citep{zheng2026say}  combines a content-code index and a discrete duration into a unified token ID. 
In contrast, \proposal makes duration part of the quantization space, so content and temporal extent jointly determine each \emph{word} in a fixed-size wordbook. 
On decoding, durations are then adjusted per occurrence rather than recovered exactly from the token ID. 
We study this representation in peripheral-to-EEG translation, where the learned segmentation is additionally coordinated with the recording intervals used for decoding.

\subsection{Parallel Decoding and Temporal Organization}

Semi-autoregressive generation reduces sequential decoding while retaining target dependencies.
\citet{wang2018semi} generate token groups autoregressively, with parallel prediction within each group.
RecoverSAT~\citep{ran2020learning} instead generates segments in parallel and tokens sequentially within each segment.
\proposal uses a related schedule.
Fixed-duration blocks are decoded in parallel, while EEG words are predicted sequentially within each block.
This keeps modality-specific segmentations aligned to a common recording timeline.

\begin{figure*}[t]
    \vspace{-2ex}
    \centering
    \includegraphics[width=0.995\textwidth,bb=0 0 791 346]{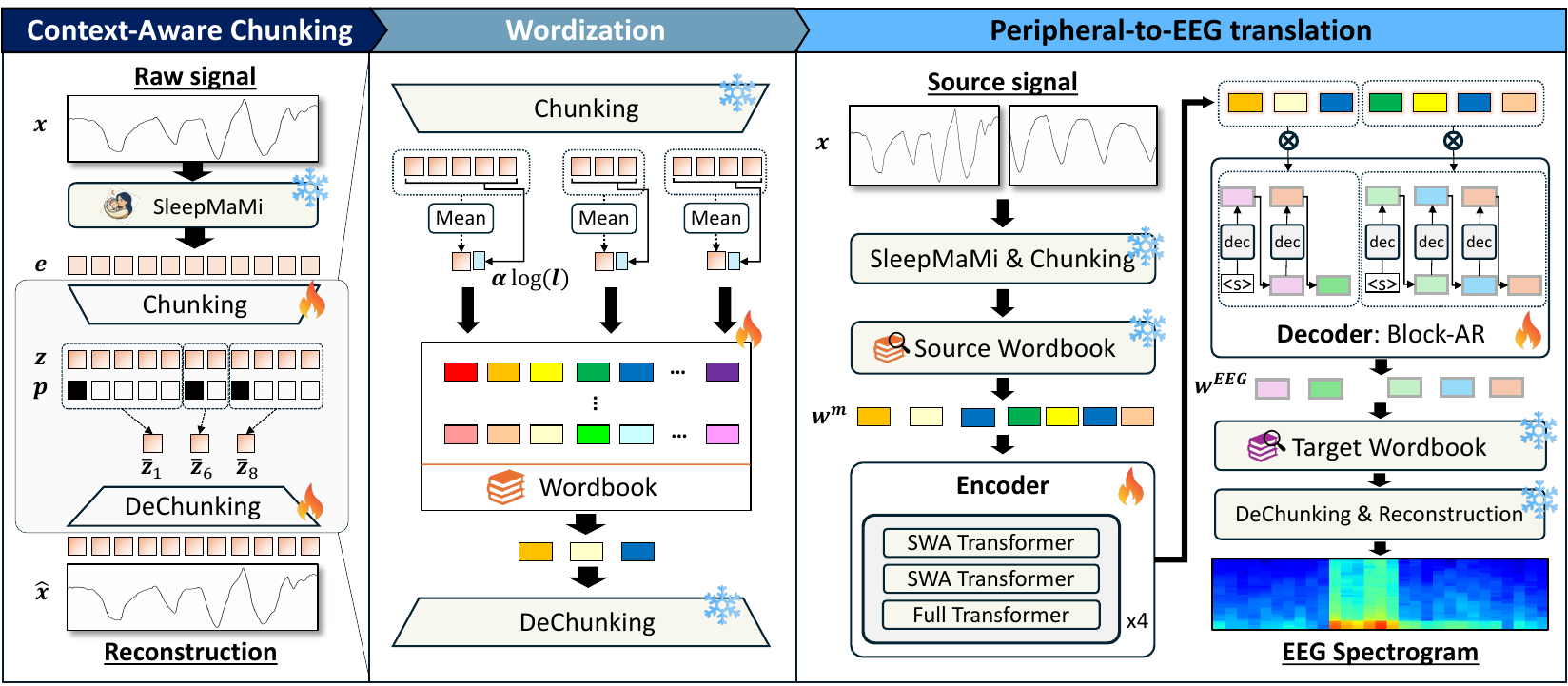}
    \vspace{-2ex}
    \caption{\textbf{Training pipeline}. \proposal is trained in three stages. The full night sequence is segmented into context-based chunks and discretized by modality-specific wordbook. The wordbook encodes both content and duration. The full night word sequence is processed by an encoder, followed by block-autoregressive decoder for temporal consistency. }
    \label{fig:method}
    \vspace{-3ex}
\end{figure*}

\section{Method} \label{sec:method}
\subsection{Overview and Problem Formulation}
Our objective is to estimate an EEG spectrogram from peripheral physiological signals. 
Let  $\mathbf{x}^m \in \mathbb{R}^L$ denote a full night recording from peripheral input $m$, where $L=f_s N_s$, $f_s$ is the sampling rate and $N_s$ is the recording duration in seconds.
We transform the paired target EEG signal into a multi-taper spectrogram~\citep{prerau2017sleep}, denoted by $\mathbf{y} \in \mathbb{R}^{T_y \times N_f}$ where $T_y$ is the number of non-overlapping $\omega$-second windows and $N_f$ is the number of frequency bins (See Appendix~\ref{sec:spectrogram} for details).
For clarity, we omit the modality superscript where the distinction is evident.

\proposal estimates the target through a compact intermediate representation. 
(i) \emph{Context-dependent chunking} first groups adjacent embeddings into variable-duration segments. 
(ii) \emph{Wordization} maps each segment to a reusable content-duration word.  
(iii) The translator then predicts EEG word identities and their occurrence-specific durations within common recording intervals, and the EEG reconstruction modules map the resulting sequence to a spectrogram. 
The overall pipeline is depicted in Figure~\ref{fig:method}.

% Figure~\ref{fig:method} provides the overview of our method. 
% Instead of translating fixed-length patch embeddings directly, our framework first compresses each modality into a stream of \emph{variable-length chunks}. It then discretizes the chunks with a learned \emph{wordbook}, and formulates peripheral-to-EEG translation as block-autoregressive prediction of the target word sequence.
% Training proceeds in three stages: (i) \emph{chunking} (ii) \emph{wordization} and (iii) \emph{translation}.

\paragraph{Base embeddings.}
We construct a modality-specific wordbooks using shared embeddings produced by the micro-encoder of SleepMaMi~\citep{park2026sleepmami}, a recent sleep foundation model.
Whereas the original SleepMaMi forms shared representations by concatenating embeddings across all modalities, we modify its shared module to process each modality independently.
Consequently, the source signal can be encoded without access to the target modality.
Further details of this modification are provided in the Appendix~\ref{sec:sleepmami}.

\subsection{Context-Dependent Chunking}
In the first stage, we learn to divide each full-night representation sequence into variable-length chunks whose boundaries depend on temporal context.
% For this purpose, a dynamic chunking modules~\citep{hwang2026dynamic} is employed.
To this end, we adapt the dynamic chunking module of~\citet{hwang2026dynamic} to physiological reconstruction.
Given a sequence of SleepMaMi embeddings $\mathbf{e} \in \mathbb{R}^{L' \times d}$, where $L'$ is the number of base patches in the full night recording and $d$ is the embedding dimension, the chunking module refines the embeddings and predicts the probability $p_i$ that base position $i$ begins a new chunk:
\begin{equation}
     (\mathbf{z}, \mathbf{p}) = \mathrm{Chunk}(\mathbf{e}), \quad \mathbf{z} \in \mathbb{R}^{L' \times d}, \quad \mathbf{p} \in [0,1]^{L'}.
\end{equation}
During the forward pass, the start positions are $\{1\} \cup\{i \ge 2 \text{ and } p_i \ge 0.5\}$. 
Let $s_1<\cdots<s_C$ denote the resulting start positions, with $s_{C+1}=L'+1$. 
The representation of the $c$-th chunk is obtained by average pooling:
\begin{equation}
    \bar{\mathbf{z}}_{c}=\frac{1}{l_c}\sum_{t=s_c}^{s_c+l_c-1}\mathbf{z}_t, \quad l_c=s_{c+1}-s_c.
\end{equation}
The pair $(s_c,l_c)$ is retained alongside each pooled representation to preserve its position and temporal length.
The chunk representations are subsequently scattered back to their original positions:
\begin{equation}
    \mathbf{u}_t=\bar{\mathbf{z}}_{c} \quad \text{for} \quad s_c \le t < s_{c+1}
\end{equation}
The resulting sequence is processed by the dechunking module and a convolutional reconstruction head, which recovers the original waveform for peripheral modalities and the corresponding spectrogram for EEG.
% Otherwise, the gradient cannot reach $\mathbf{p}$ because hard boundary condition is used.
We use mean squared error as reconstruction objective ($\mathcal{L}_{\mathrm{recon}}$). 

Because hard thresholding prevents reconstruction gradients from propagating through the boundary decisions, we apply a straight-through surrogate by replacing  $\mathbf{u}_t$ with $\tilde{\mathbf{u}}_t=\mathbf{u}_t (1 + (p_{s_c}-\mathrm{sg}[p_{s_c}]))$ where $\text{sg}[\cdot]$ denotes the stop-gradient operator. 
This does not influence the forward pass, but allows the gradient to reach $\mathbf{p}$ during backpropagation. 
% \begin{equation}
%     \mathcal{L}_{recon} = 
%          \begin{cases}
%             \frac{1}{T_y}\sum_{v}{\|\textbf{y}_v - \hat{\textbf{y}}_v\|_2^2} & \quad m=EEG \\
%             \frac{1}{L}\sum_{t}{\|x_t - \hat{x}_t\|_2^2} & \quad otherwise            
%          \end{cases} \\
% \end{equation}

In addition to the reconstruction objective, we introduce three auxiliary losses. 
First, following ~\citep{hwang2026dynamic}, we adopt a ratio loss ($\mathcal{L}_{\mathrm{ratio}}$) to regulate the proportion of predicted boundaries. 
Second, an alignment loss ($\mathcal{L}_{\mathrm{align}}$) anchors the refined embeddings to the original shared embeddings, preventing them from deviating substantially toward modality-specific representation space. 
Finally, the barrier loss ($\mathcal{L}_{\mathrm{barrier}}$) encourages boundaries to start at the common recording intervals ($B$ base positions).
This constraint promotes periodic synchronization between the source and target sequences, facilitating the block autoregressive translation described in Section~\ref{sec:translation}. 
The details of these auxiliary losses are provided in Appendix~\ref{sec:losses}. 

The overall chunking objective is defined as follows:
\begin{equation}
    \begin{aligned}
         \mathcal{L}_{\mathrm{chunk}} &= \mathcal{L}_{\mathrm{recon}}  + \lambda_{\mathrm{ratio}}\mathcal{L}_{\mathrm{ratio}} + \lambda_{\mathrm{align}}\mathcal{L}_{\mathrm{align}} + \lambda_{\mathrm{barrier}}\mathcal{L}_{\mathrm{barrier}} \\    
     \end{aligned}
\end{equation}

\subsection{Wordization} 
% Pooling produces one vector for each segment, but does not specify its temporal extent. 
We construct a modality-specific wordbook $\mathcal{W}=\{w_k\}_{k=1}^K$ from the learned chunk representations.
To incorporate temporal extent into each word, we augment the pooled representation with log duration: 
$\mathbf{z}'_{c} = [\bar{\mathbf{z}}_{s_c}; \alpha \text{log}(l_c)]$ where $\alpha$ controls the scale of duration relative to content.
We represent each wordbook entry as $w_k=[\nu_k;\eta_k]$, where $\nu_k\in\mathbb{R}^d$ represents its content and $\eta_k \in \mathbb{R}$ represents its duration component. Each chunk $\mathbf{z}'_{c}$ is assigned to its nearest wordbook entry following this rule:
\begin{equation}
    k_c = \arg\min_{k \in \{1,...,K\}}[\|\mathbf{\nu}_k - \bar{\mathbf{z}}_{c}\|_2^2+(\alpha \log l_c - \eta_k)^2] %, \quad \mathbf{q}_c=\mathbf{w}_{k_c}
\end{equation}
where $k_c$ is the discrete word index. % and $\mathbf{q}_c$ is its corresponding wordbook embedding.
This formulation helps both content and duration to contribute directly to word identity.
In addition, for each word, we maintain an exponential moving average $\mu_k$ of the durations of its assigned chunks for use in subsequent duration prediction.

During wordization, the chunking module is frozen. Each chunk is assigned to a word index $k_c$ and the corresponding word embedding is passed through the dechunking module and reconstruction head to recover the original peripheral signal or EEG spectrogram.
The wordization objective is 
\begin{equation}
    \mathcal{L}_{\mathrm{Wordization}} = \mathcal{L}_{recon} + \frac{\lambda_Q}{C}\sum_{c=1}^{C}\|\mathrm{sg}[\mathbf{z}'_{c}]  -  \mathbf{w}_{k_c}\|_2^2
\end{equation}
With fixed chunk representations and fixed $\alpha$, we do not include the term $\|\mathbf{z}'_{c}  -  \text{sg}[\mathbf{w}_{k_c}]\|_2^2$, unlike conventional vector quantization~\citep{van2017neural}.

We initialize the wordbook and calculate $\alpha$ using statistics from the first $I$ batches (We use $I=10$) produced by the frozen chunking module.
The calibration of $\alpha$ balances the contributions of content and log-duration to word assignment, encouraging sensitivity to the chunk duration without requiring a distinct word for every possible length. More details of wordbook initialization are provided in Appendix~\ref{sec:implementation}.

% In the second stage, we discretize the dynamically segmented representations into a fixed-size wordbook $\mathcal{W} = \{\mathbf{w}_k\}_{k=1}^K$ containing $K$ entries. 
% Because chunk duration carries physiologically relevant information, we append the log-duration to each pooled chunk representation:

% \begin{equation}
%     \mathbf{z}'_{s_c} = [\bar{\mathbf{z}}_{s_c}; \alpha \cdot \text{log}(l_c)] \in \mathbb{R}^{d+1}
% \end{equation}
% where $\alpha$ controls the relative contribution of duration. 
% Accordingly, each wordbook entry is also defined in $\mathbb{R}^{d+1}$. 

\subsection{Peripheral to EEG translation} \label{sec:translation}
After constructing separate wordbooks for the source and target modalities, we train an encoder-decoder model to translate a sequence of peripheral words into a sequence of EEG words, with chunking, dechunking and reconstruction modules frozen.
Before target generation begins, the encoder processes the complete peripheral word sequence $\{\mathbf{w}_{k_c}^m\}_{c=1}^{C_m}$.
Rotary positional embeddings~\cite{su2024roformer} encode the original start position of each source word, retaining recording time information despite variable word durations.
The encoder consists of four Transformer blocks~\cite{vaswani2017attention}, each containing two sliding window self-attention layers~\citep{beltagy2020longformer} followed by one full-attention layer.
This architecture captures both local temporal patterns and full-night source context.

The decoder generates target EEG words conditioned on the encoded source sequence.
Because the source and target modalities are chunked independently, their word sequences may differ in both the number and duration of their words.
Na\"ive word-to-word decoding therefore accumulates temporal misalignment over the course of a full-night recording.
% In contrast to natural language translation, where differences in output length are generally permissible, physiological signal translation requires close temporal correspondence between the source and target sequences. 
To constrain this drift, we employ \textbf{Block AutoRegressive (AR)} decoding. 
Let $\mathbf{H}^m_b$ denote the encoded source memory associated with the $b$-th block.
Within each block, the decoder generates EEG words autoregressively while conditioning on $\mathbf{H}^m_b$. 
% Different blocks can therefore be decoded in parallel. 
This formulation bounds any temporal discrepancy within individual blocks and reduces the sequential decoding path relative to full-night AR generation.
% It also improves the inference efficiency by enabling word generation to be parallelized across blocks rather than performed autoregressively over the entire night-long sequence.
% The barrier loss ($\mathcal{L}_{\mathrm{barrier}}$), introduced during chunking encourages word boundaries to align with block boundaries, thereby preventing individual words from spanning adjacent blocks.

Let $N_B=\left\lceil L'/B \right\rceil$ be the number of blocks and $n_b$ the number of target words in block $b$.
For training labels $k_{b,s}^{EEG}$ obtained from frozen EEG tokenizer, the word-identity objective is
\begin{equation}
 \mathcal{L}_{\mathrm{id}} = -\frac{1}{N_B}\sum_{b=1}^{N_B}\frac{1}{n_b}\sum_{s=1}^{n_b}\log p(k_{b,s}^{EEG}|k_{b,<s}^{EEG}, \mathbf{H}^m_b) 
\end{equation}

\textbf{Occurrence-specific duration}. In addition to word identity, the decoder predicts a duration residual $\hat{r}_{b,s}$ relative to the EMA of the selected target word.
Each word represents a prototypical content-duration pattern, and the residual accommodates variation across individual occurrences.
Using target tokenizer residual $r_{b,s}=l_{b,s}^{EEG} - \mathrm{round}(\mu_{k_{b,s}})$ as supervision, the duration and  overall translation loss are defined as
\begin{equation}
    \mathcal{L}_{\mathrm{dur}} = -\frac{1}{N_B}\sum_{b=1}^{N_B}\frac{1}{n_b}\sum_{s=1}^{n_b}\log p(r_{b,s}^{EEG}|k_{b,s}^{EEG}, \mathbf{H}^m_b), \quad \mathcal{L}_{\mathrm{trans}} =  \mathcal{L}_{\mathrm{id}} + \lambda_{\mathrm{dur}}\mathcal{L}_{\mathrm{dur}}
\end{equation}

\textbf{Training}. Throughout translation training, the chunking, dechunking and reconstruction modules remain frozen.
Training begins with teacher-forcing after which scheduled sampling~\citep{bengio2015scheduled} is applied with  a progressively increasing probability to mitigate exposure bias.

\textbf{Inference and reconstruction}. 
During inference, the target words are generated within each block until their cumulative predicted duration reaches the corresponding block length in base positions.
For block $b$, this target length is $T_b=\min\left(B, L'-(b-1)B\right)$.
If the cumulative duration exceeds $T_b$, the duration of the final word is truncated.
If the generated sequence length falls short, the duration of the final word is extended to fill the remaining positions.
The predicted EEG word embeddings are then expanded onto the base grid using the predicted durations and processed by the frozen EEG dechunker and reconstruction head to obtain the full night EEG spectrogram.

% During training, each target segment supplies a length $l_{s}^{EEG}$ an EEG word label $k_{s}^{EEG}$ and  from the frozen target chunking module.

% The decoder predicts a residual adjustment to this entry-specific average, and their sum determines the final target word duration.
% \begin{equation}
%     \hat{l}_s = \mathrm{EMA}(l_s) + \mathrm{Linear}(\hat{\mathbf{w}}_{s}^{EEG})
% \end{equation}

\section{Experiments} \label{sec:experiments}
\subsection{Datasets, Preprocessing and Baselines}
We train on a combined set of nine publicly available sleep datasets: the Sleep Heart Health Study (SHHS1/2)~\citep{zhang2018national, quan1997sleep}, Korea Image-based Sleep Study (KISS)~\citep{jeong2023standardized}, Korea Video Sleep Study (KVSS)~\citep{choi2026non}, Physionet 2018 (PHY)~\citep{goldberger2000physiobank, ghassemi2018you}, Multi-Ethnic Study of Atherosclerosis (MESA)~\citep{chen2015racial} and Osteoporotic Fractures in Men Study (MrOS)~\citep{blackwell2011associations}. 
% The SHHS1 and MESA are partitioned into training, validation and test sets according to established protocols~\citep{phan2021xsleepnet, park2025distillsleep}. 
Overall, we use 20,036 PSG recordings (107,469 hours) for training. 
We use the Cleveland Family Study (CFS)~\citep{zhang2018national, redline1995familial} as a held-out dataset to assess \proposal's generalizability to previously unseen data.  Further dataset details are provided in Appendix~\ref{sec:datasets}.

We apply standard preprocessing to all modalities. 
EEG signals are filtered using a 0.3–40 Hz bandpass filter, and ECG signals undergo a 0.3–60 Hz bandpass filtering. 
A 60 Hz notch filter is additionally applied to both EEG and ECG signals.
Respiratory signals are low-pass filtered at 15 Hz.
Biosignals are resampled at 100 Hz to harmonize the differing sampling rates across the diverse datasets. 
We use C4-A1 EEG channel and abdominal belt respiratory signal.
For PPG, the raw signal is polarity-corrected, low-pass filtered at 8 Hz and clipped at $\pm3$ standard deviations.
Before model input, all signals are z-score normalized independently within each 30-second epoch.

As a baseline, we construct a simple CNN-Transformer model for peripheral-to-EEG conversion (referred to C-T baseline, details in  Appendix~\ref{sec:model}).
For respiratory-to-EEG conversion, we compare against the recently proposed Physiology as Language (PasL)~\citep{zha2026physiology}. We reproduce PasL following the paper, since the source code is not released.
wav2sleep~\citep{carter2025wav}, a sleep staging specialized model using peripheral inputs, is also adopted to compare against raw signal based method.

\textbf{Baseline split alignment.}
For SHHS1, MESA and CFS, we align our evaluation sets with the official wav2sleep splits to ensure comparisons are performed on the same recordings. Details of the split construction are provided in Appendix~\ref{sec:datasets}.

\subsection{EEG Reconstruction Performance}
Table~\ref{tab:recon_performance} provides the EEG spectrogram reconstruction quality of \proposal and other baselines. Following PasL, we report the quality using two metrics: Mean Absolute Error (MAE) and Signal-to-Noise Ratio (SNR), in comparison to the ground truth spectrogram. Across all source modality inputs and datasets, \proposal exhibits lower MAE and higher SNR compared to other baselines. Particularly, \proposal demonstrates its robustness to the unseen dataset (CFS), while PasL shows suboptimal performance.

\begin{table}[t]
\centering
\caption{\textbf{EEG spectrogram reconstruction performance}. MAE and SNR are provided across modalities and datasets. Bold values indicate the best among spectrogram-generation methods.}
\vspace{-2ex}
\label{tab:recon_performance}
\resizebox{0.9\linewidth}{!}{%
\begin{tabular}{ccccc cc cc}
\toprule
\multirow{2}{*}{\textbf{Dataset}} & \multirow{2}{*}{\textbf{Metric}}
& \multicolumn{3}{c}{\textbf{Respiratory}}
& \multicolumn{2}{c}{\textbf{ECG}}
& \multicolumn{2}{c}{\textbf{PPG}} \\
\cmidrule(lr){3-5} \cmidrule(lr){6-7} \cmidrule(lr){8-9}
& & Ours & PasL & C-T baseline & Ours & C-T baseline & Ours & C-T baseline\\
\midrule
% \multirow{2}{*}{SHHS1 (Internal)}
SHHS1                   & MAE ($\downarrow$)    & \textbf{0.047} & 0.074 & 0.050 & \textbf{0.048} & 0.056 & - & - \\
\small{(Internal)}      & SNR ($\uparrow$)      & \textbf{17.35} & 16.03 & 16.86 & \textbf{17.26} & 15.71 & - & - \\
\midrule
MESA        & MAE ($\downarrow$)                & \textbf{0.056} & 0.069 & 0.064 & \textbf{0.059} & 0.075 & \textbf{0.057} & 0.066 \\
\small{(Internal)}  & SNR ($\uparrow$)          & \textbf{14.38} & 11.74 & 13.88 & \textbf{14.12} & 12.51 & \textbf{14.23} & 12.93 \\
\midrule
CFS         & MAE ($\downarrow$)                & \textbf{0.058} & 0.112 & 0.061 & \textbf{0.065} & 0.067 & \textbf{0.059} & 0.060 \\
\small{(External)}  & SNR ($\uparrow$)          & \textbf{15.26} & 13.89 & 14.12 & \textbf{14.38} & 13.71 & \textbf{14.69} & 14.67 \\
\bottomrule
\end{tabular}
}
\vspace{-1ex}
\end{table}

\begin{table}[t]
\captionsetup{justification=raggedright,singlelinecheck=false}
\centering
\caption{\textbf{Sleep staging performance}. Best spectrogram-generation results are bolded.}
\vspace{-2ex}
\label{tab:sleep_staging}
\resizebox{0.99\linewidth}{!}{%
\begin{tabular}{ccc cccc ccc ccc}
\toprule
\multirow{3}{*}{\textbf{Dataset}} &
\multirow{3}{*}{\textbf{Metric}} &
\textbf{EEG} &
\multicolumn{4}{c}{\textbf{Respiratory}} &
\multicolumn{3}{c}{\textbf{ECG}} &
\multicolumn{3}{c}{\textbf{PPG}} \\
\cmidrule(lr){3-3} \cmidrule(lr){4-7} \cmidrule(lr){8-10} \cmidrule(lr){11-13}
& & GT & \multicolumn{3}{c}{Spectrogram generation} & Specialized
& \multicolumn{2}{c}{Spectrogram generation} & Specialized
& \multicolumn{2}{c}{Spectrogram generation} & Specialized \\
\cmidrule(lr){4-6}  \cmidrule(lr){7-7} \cmidrule(lr){8-9}  \cmidrule(lr){10-10}\cmidrule(lr){11-12} \cmidrule(lr){13-13}

& & Spec & \proposal & PasL & C-T baseline & wav2sleep
& \proposal & C-T baseline & wav2sleep
& \proposal & C-T baseline & wav2sleep \\

\midrule
\multirow{3}{*}{SHHS1}
& MF1       & 0.872 & \textbf{0.757} & 0.668 & 0.640 & 0.777 & \textbf{0.714} & 0.671 & 0.788 & - & - & -\\
& Acc.      & 0.879 & \textbf{0.766} & 0.681 & 0.661 & 0.806 & \textbf{0.729} & 0.689 & 0.814 & - & - & - \\
& Kappa     & 0.822 & \textbf{0.658} & 0.531 & 0.503 & 0.711 & \textbf{0.603} & 0.547 & 0.724 & - & - & - \\
\midrule
\multirow{3}{*}{MESA}
& MF1       & 0.858 & \textbf{0.757} & 0.642 & 0.677 & 0.757 & \textbf{0.699} & 0.583 & 0.753 & \textbf{0.709} & 0.332 & 0.746 \\
& Acc.      & 0.884 & \textbf{0.799} & 0.701 & 0.723 & 0.854 & \textbf{0.741} & 0.657 & 0.841 & \textbf{0.754} & 0.566 & 0.840 \\
& Kappa     & 0.825 & \textbf{0.694} & 0.534 & 0.584 & 0.770 & \textbf{0.609} & 0.476 & 0.750 & \textbf{0.626} & 0.216 & 0.749 \\
\midrule
\multirow{3}{*}{CFS}
& MF1       & 0.862 & \textbf{0.743} & 0.417 & 0.642 & 0.816 & \textbf{0.670} & 0.651 & 0.841 & \textbf{0.679} & 0.282 & 0.803 \\
& Acc.      & 0.866 & \textbf{0.748} & 0.563 & 0.660 & 0.834 & \textbf{0.686} & 0.662 & 0.858 & \textbf{0.688} & 0.523 & 0.821\\
& Kappa     & 0.806 & \textbf{0.639} & 0.304 & 0.511 & 0.759 & \textbf{0.546} & 0.521 & 0.794 & \textbf{0.550} & 0.154 & 0.741 \\
\bottomrule
\end{tabular}
}
\vspace{-3ex}
\end{table}

\subsection{Downstream Performance}
We evaluate whether the generated EEG spectrograms preserve information relevant to three downstream tasks: four-stage sleep staging, arousal detection, and survival prediction.

\textbf{Sleep staging}. Table~\ref{tab:sleep_staging} reports performance on four-stage sleep staging (Wake, N1/N2, N3, and REM) using Macro-F1, accuracy, and Cohen's kappa. 
For spectrogram-generation methods (\proposal, PasL, and the C-T baseline), we froze the generation model and separately fine-tuned CNN classifier on spectrograms from the training split and evaluate it on the corresponding test split of each dataset. 
For ground-truth (GT) EEG spectrograms, a CNN classifier is also fine-tuned on training split and evaluated.
For wav2sleep, we directly evaluate the released pretrained models.

Among the evaluated spectrogram-generation methods, \proposal achieves the strongest sleep staging performance across the datasets and source modalities. 
Within \proposal, respiration consistently yields better performance than ECG or PPG, suggesting that sleep-stage-related EEG structure can be recovered more reliably from respiratory signals in the evaluated cohorts.

Nonetheless, \proposal generally underperforms wav2sleep and the classifier trained on GT EEG spectrograms. 
This comparison should be interpreted in light of their different objectives: wav2sleep is optimized specifically for sleep staging using direct stage supervision, whereas \proposal learns task-agnostic cross-modal generation and is evaluated using a separate downstream classifier. 
The remaining gap to GT EEG further suggests that some stage-relevant information is either unavailable in peripheral signals or lost during word-based generation. 
Despite these differences, the consistent improvement over other spectrogram-generation baselines demonstrates that \proposal preserves more stage-discriminative EEG structure without task-specific optimization.

% \subsection{Downstream Performance}
% We test three downstream tasks using the generated EEG spectrogram: 4-stage sleep staging, arousal detection and disease prediction. 

% \textbf{Sleep staging}. Table~\ref{tab:sleep_staging} provides the 4-stage sleep staging (Wake, N1/N2, N3, REM) performance of \proposal and other models, using three metrics (Macro-F1, Accuracy and Cohen's kappa).
% CNN based classifier is finetuned on the generated spectrogram of \proposal, PasL and C-T baseline on the trainig split and evaluated on test split of each dataset. 
% Ground truth spectrogram is also evaluated using the same classifier.
% For wav2sleep, we directly use the released pretrained weights.
% Amongst spectrogram based methods, \proposal exhibits the highest performance across all datasets and modalities. 
% Comparing modalities, respiratory-to-EEG translation achieves the highest sleep staging performance across all datasets while this trend is not evident for C-T baseline or PasL. 
% There is still clear gap from wav2sleep, a sleep staging specialized method or GT spectrogram based staging results.

\textbf{Arousal detection}. 
Table~\ref{tab:arousal} compares the arousal detection performance of \proposal with those of the baseline models. 
To match the temporal resolution of the input spectrograms, we convert the second-level arousal annotations into epoch-level labels. 
An epoch is labeled as positive if it contains at least three seconds of annotated arousal activity. 
Using the epoch-level arousal labels, we train a linear classifier on the generated spectrogram.
The arousal prediction AUROC conditioned on non-wake epochs is reported, to align with the arousal determination rule~\citep{berry2012aasm}.

\proposal achieves the highest AUROC among spectrogram-generation methods across datasets and modalities.
In particular, it consistently outperforms wav2sleep.
Since wav2sleep is trained for sleep staging, this comparison indicates that its representations carry little arousal information beyond what staging requires.
This supports the use of generated spectrograms as a task-agnostic intermediate rather than a proxy for a single label.

\begin{wrapfigure}{b}{0.4\textwidth}
    \vspace{-6ex}
    \centering
    \includegraphics[width=\linewidth,bb=0 0 288 216]{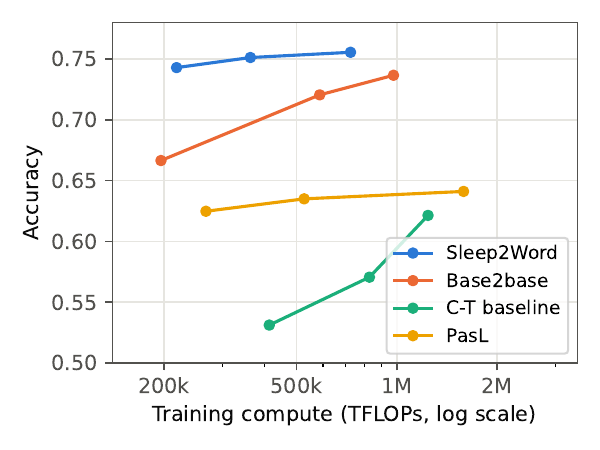}
    \vspace{-5ex}
    \caption{\textbf{Compute efficiency.} Sleep staging accuracy versus cumulative training compute. \proposal achieves the best accuracy-compute trade-off among the compared methods}
    \label{fig:compute_efficiency}
    \vspace{-8.5ex}
\end{wrapfigure}

\textbf{Survival prediction}. 
We evaluate prognostic information in generated EEG spectrograms on SHHS1 using a linear Cox head for all-cause mortality, CVD mortality, and stroke.
\proposal achieves higher C-indices than PasL and C-T across all respiratory-input outcomes, and exceeds wav2sleep for CVD mortality while remaining comparable for the other outcomes.
With ECG input, \proposal achieves higher C-indices than C-T and wav2sleep across all outcomes (Table~\ref{tab:survival}).

%%%%%%%%%%%%%%%%%%

\begin{table}[t]
\centering
\caption{\textbf{Arousal detection performance}. Results are provided in AUROC conditioned on non-wake epochs. Bold values indicate the best performance among spectrogram-generation methods. }
\vspace{-2ex}
\label{tab:arousal}
\resizebox{0.99\linewidth}{!}{%
\begin{tabular}{cc cccc ccc ccc}
\toprule
\multirow{2}{*}{\textbf{Dataset}} &
\textbf{EEG} &
\multicolumn{4}{c}{\textbf{Respiratory}} &
\multicolumn{3}{c}{\textbf{ECG}} &
\multicolumn{3}{c}{\textbf{PPG}} \\
\cmidrule(lr){2-2} \cmidrule(lr){3-6} \cmidrule(lr){7-9} \cmidrule(lr){10-12}
& GT & \multicolumn{3}{c}{Spectrogram generation} & Specialized
& \multicolumn{2}{c}{Spectrogram generation} & Specialized
& \multicolumn{2}{c}{Spectrogram generation} & Specialized \\
\cmidrule(lr){3-5}  \cmidrule(lr){6-6} \cmidrule(lr){7-8}  \cmidrule(lr){9-9}\cmidrule(lr){10-11} \cmidrule(lr){12-12}

& Spec
& \proposal & PasL & C-T baseline & wav2sleep
& \proposal & C-T baseline & wav2sleep
& \proposal & C-T baseline & wav2sleep \\
\midrule
SHHS1 & 0.879 & \textbf{0.799} & 0.669 & 0.743 & 0.666 & \textbf{0.802} & 0.728 & 0.681 & - & - & - \\
MESA  & 0.827 & \textbf{0.824} & 0.633 & 0.760 & 0.655 & \textbf{0.779} & 0.662 & 0.650 & \textbf{0.781} & 0.521 & 0.647 \\
CFS   & 0.891 & \textbf{0.777} & 0.660 & 0.772 & 0.595 & \textbf{0.788} & 0.721 & 0.616 & \textbf{0.757} & 0.511 & 0.588 \\
\bottomrule
\end{tabular}%
}
\vspace{-2ex}
\end{table}

% Sleep staging
% Arousal detection

%%%%%%%%%%%%%%%%%% Survival analysis: all-cause, cvd death, stroke outcomes from SHHS1 %%%%%%%%%%%%%%%%%%
\begin{table}[!t]
\centering
\caption{\textbf{Survival prediction performance on SHHS1}. Results are reported using the C-index. Bold values indicate the best performance among spectrogram-generation methods.}
\vspace{-2ex}
\label{tab:survival}
\resizebox{0.99\linewidth}{!}{%
\begin{tabular}{cc cccc ccc}
\toprule
\multirow{3}{*}{\textbf{Dataset}} &
\multirow{3}{*}{\textbf{Outcome}} &
\multicolumn{4}{c}{\textbf{Respiratory}} &
\multicolumn{3}{c}{\textbf{ECG}} \\

\cmidrule(lr){3-6} \cmidrule(lr){7-9}

& &
\multicolumn{3}{c}{Spectrogram generation} & Specialized &
\multicolumn{2}{c}{Spectrogram generation} & Specialized \\

\cmidrule(lr){3-5} \cmidrule(lr){6-6}
\cmidrule(lr){7-8} \cmidrule(lr){9-9}

& &
\proposal & PasL & C-T baseline & wav2sleep &
\proposal & C-T baseline & wav2sleep \\

\midrule

\multirow{3}{*}{SHHS1}
& All-cause mortality
& \textbf{0.700} & 0.645 & 0.574 & 0.697
& \textbf{0.736} & 0.680 & 0.713 \\

& CVD mortality
& \textbf{0.832} & 0.648 & 0.596 & 0.758
& \textbf{0.826} & 0.729 & 0.775 \\

& Stroke
& \textbf{0.824} & 0.618 & 0.587 & 0.824
& \textbf{0.837} & 0.833 & 0.820 \\

\bottomrule
\end{tabular}%
}
\vspace{-3ex}
\end{table}
%%%%%%%%%%%%%%%%%% Survival analysis: neurodegenrative diseases from SHHS1 %%%%%%%%%%%%%%%%%%

\subsection{Compute efficiency}
To assess the computational efficiency of \proposal, we compare its cumulative training compute, measured in TFLOPs, with linear probed sleep staging accuracy on SHHS1.
Here, we introduce Base2base, which directly translates a sequence of source-modality base patch embeddings $\mathbf{e}^m$, into the corresponding sequence of EEG base patch embeddings $\mathbf{e}^{EEG}$, without intermediate chunking or wordization.
As shown in Figure~\ref{fig:compute_efficiency}, \proposal achieves higher accuracy than the baselines at comparable or lower compute budgets. This result demonstrates the superior accuracy-compute trade-off enabled by the word-based framework.
\vspace{-2ex}

\subsection{Qualitative Analysis and Ablation}
\begin{figure*}[t]
    \centering
    \begin{subfigure}[t]{0.48\textwidth}
        \centering
        \includegraphics[width=\linewidth]{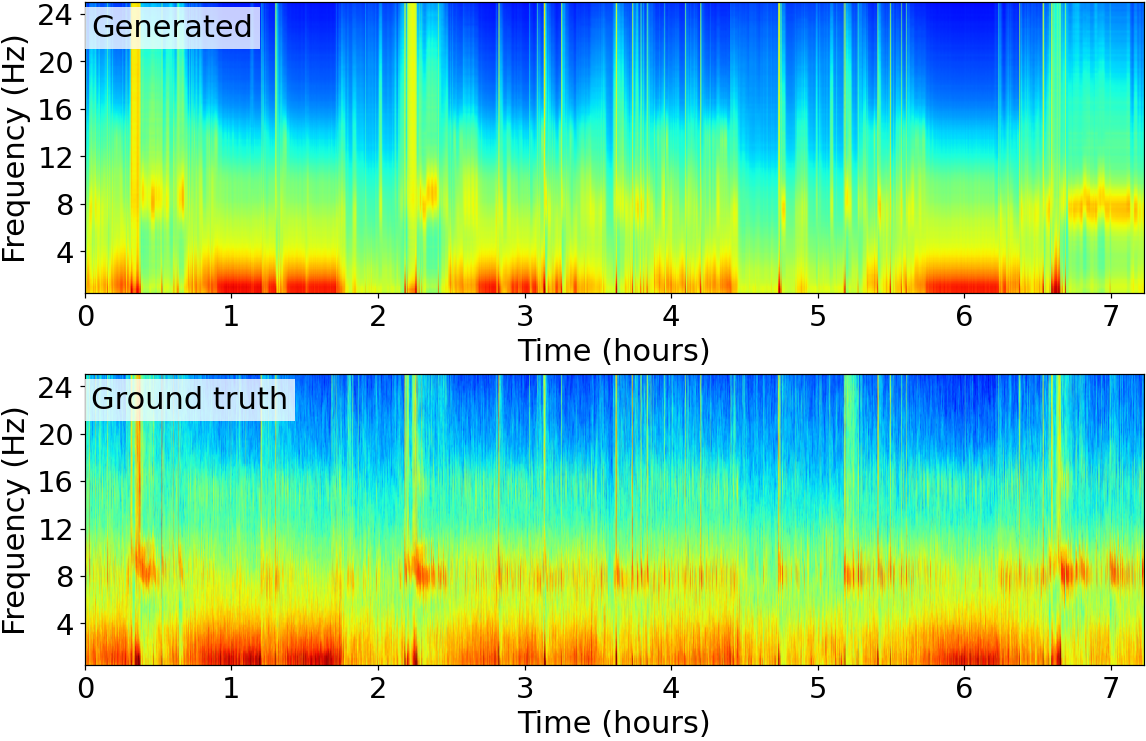}
        \vspace{-3ex}
        \caption{Spectrogram (from Resp.): 60-year-old Male}        
        \label{fig:spec_sample1}
    \end{subfigure}
    \hfill
    \begin{subfigure}[t]{0.48\textwidth}
        \centering
        \includegraphics[width=\linewidth]{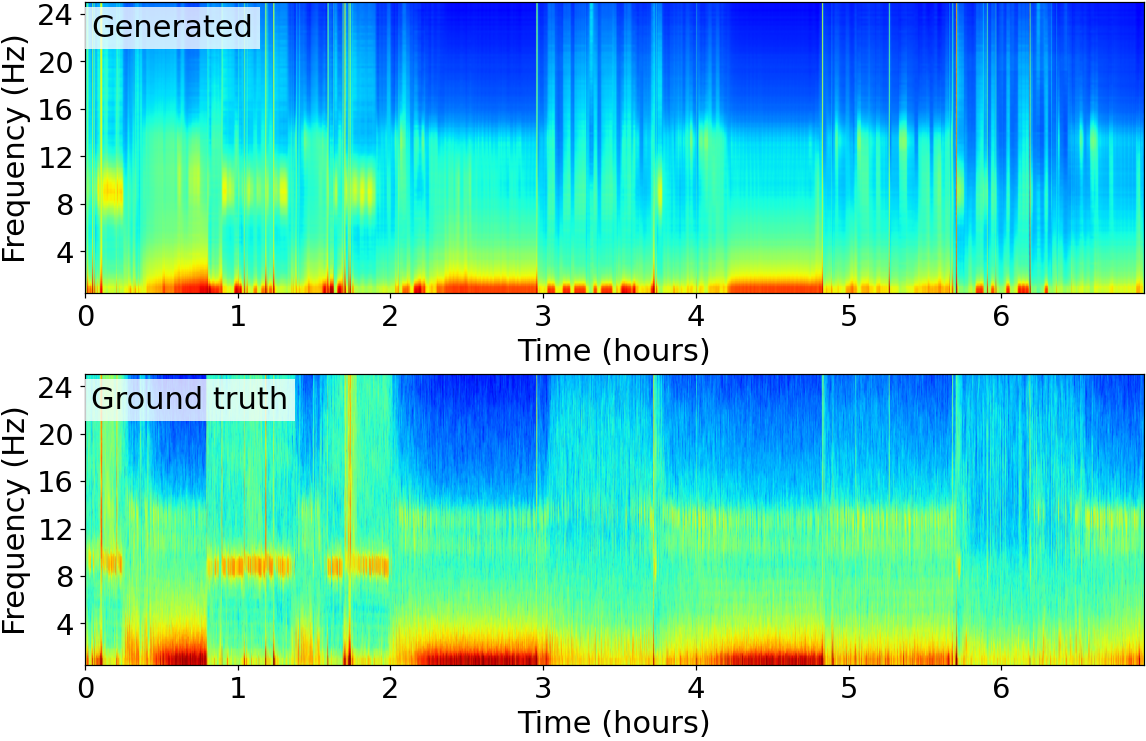}
        \vspace{-3ex}
        \caption{Spectrogram (from ECG): 54-year-old Male}        
        \label{fig:spec_sample2}
    \end{subfigure}
    
    \begin{subfigure}[t]{0.48\textwidth}
        \centering
        \includegraphics[width=\linewidth]{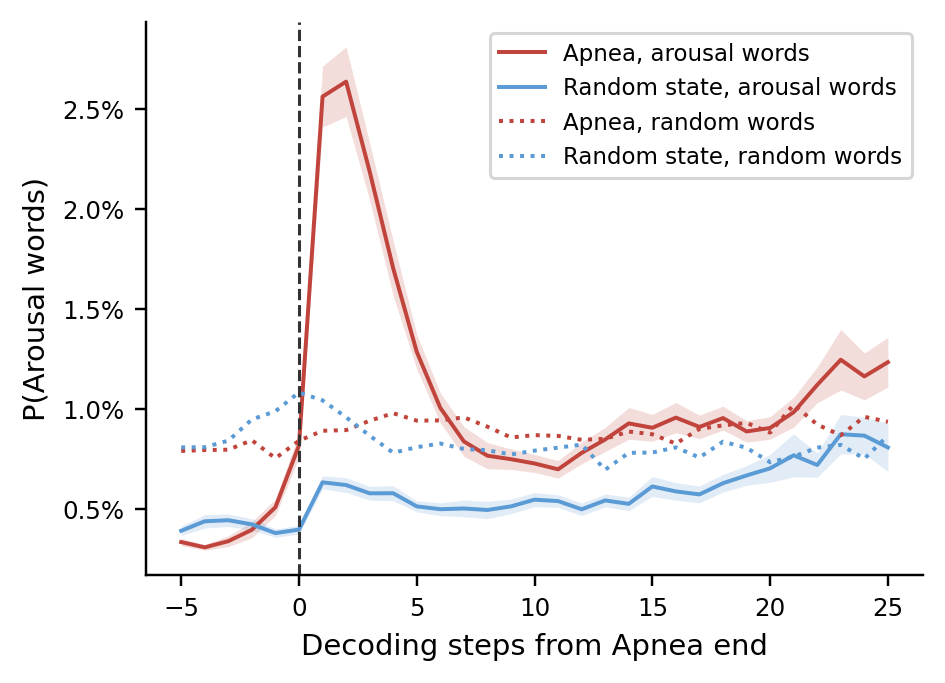}
        \vspace{-2ex}
        \caption{EEG arousal-word probabilities on apnea input}
        \vspace{-1ex}
        \label{fig:arousal_word_probability}        
    \end{subfigure}
    \hfill
    \begin{subfigure}[t]{0.48\textwidth}
        \centering
        \includegraphics[width=\linewidth]{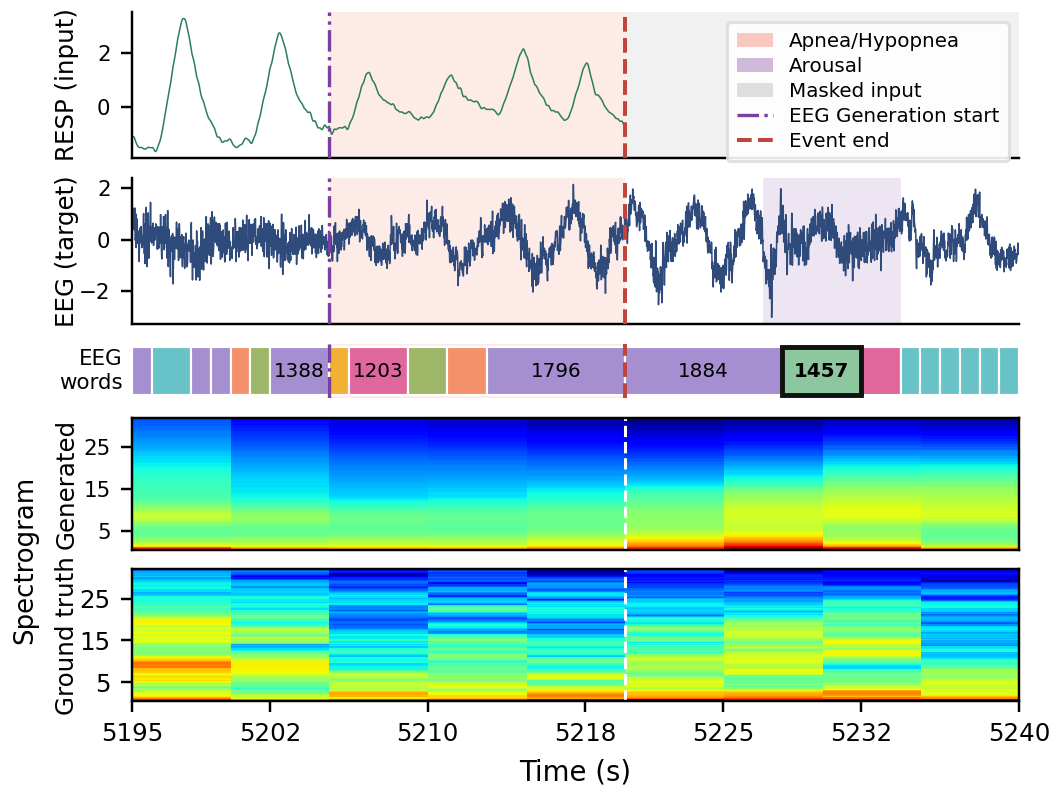}
        \vspace{-2ex}
        \caption{Arousal generation example after apnea}
        \vspace{-1ex}
        \label{fig:apnea_case_study}
    \end{subfigure}

    \caption{\textbf{Qualitative analysis}.
    (a-b) Full-night comparison between the generated and ground-truth EEG
    spectrograms.
    (c) Changes in the probability of generating arousal-related EEG words
    after apnea termination.
    (d) An example showing the input respiratory signal from respiratory event,
    generated EEG words, and spectrogram. The bolded box represents an arousal word.}
    \label{fig:qualitative_analysis}
\end{figure*}

Figures~\ref{fig:spec_sample1} and~\ref{fig:spec_sample2} compare representative EEG spectrograms generated from peripheral signals with their corresponding ground-truth spectrograms. The generated spectrograms reproduce broad full-night spectral structure, including variations in low frequency, spindle frequency range, and higher frequency power, although they do not capture all fine-grained structures present in the ground-truth EEG.
To examine whether the model captures cross-modal physiological relationships, we investigate apnea-arousal coupling.
We identify the 20 EEG words (the top-1\% of wordbook) most strongly associated with annotated arousals and provide respiratory input only until apnea termination, while allowing EEG generation to continue beyond the event. 
Figure~\ref{fig:arousal_word_probability} shows that the probability of arousal-related EEG words increases sharply after apnea termination and exceeds those under random-state conditioning. 
The example in Figure~\ref{fig:apnea_case_study} further illustrates this pattern, suggesting that the model learns the temporal relationship between disordered breathing and subsequent cortical arousal.
Together, these results highlight the potential of word-based approach for model-based exploration of cross-modal physiological associations.
% This case study shows the potential use case of \proposal that could be used to understand the physiological relationship between modalities in a model-driven manner.

Ablation results (Appendix~\ref{sec:ablation}) confirm the contribution of each core component: loss components (Table~\ref{tab:ablation_loss}), chunking and wordization (Table~\ref{tab:ablation_chunking}), duration modeling (Table~\ref{tab:ablation_duration}), wordbook size (Table~\ref{tab:ablation_K}), target chunking ratio (Table~\ref{tab:ablation_N}) and block size (Table~\ref{tab:ablation_B}).
\vspace{-2ex}

% We also verify whether the model learns physiological grammar across the modality.
% TO do so, we test whether the model predicts the EEG arousal when it is given the disordered breathing signal.
% The EEG arousal words are first determined as the top 20 EEG words that overlaps the arousal annotated period. 
% Then we give the model respiratory signal from apnea period and make the model generate EEG words beyond the apnea period. 
% Note that the source input is only given until apnea ends, so the model cannot translate the arousal respiratory words into arousal EEG words, but it should infer the arousal commonly comes after the apnea period.
% Figure~\ref{fig:arousal_word_probability} shows the probabilities of arousal EEG words when respiratory apnea signal is provided. 
% The probabilities of arousal EEG words showed clearly higher value after the apnea ends, and compared to random state. 
% Figure~\ref{fig:apnea_case_study} shows an example of this apnea-arousal generation. 
% This verifies that the model not only translates input sequence to output sequence, but also has basic understanding of physiology between modalities.

% Wordbook size
% Number of chunks
% Block size
\section{Conclusion}
In this paper, we present \proposal, a word-based framework for peripheral-to-EEG translation. 
By introducing context-dependent chunking, duration-aware physiological wordization, and a block-AR decoder, our model captures the distinct temporal structures of different modalities while effectively mitigating temporal drift. 
Our evaluations confirm that \proposal outperforms existing baselines in EEG spectrogram reconstruction and clinical downstream tasks. 
Finally, the word-based architecture reduces the sequence length by 69.7\%, enabling a favorable compute-accuracy trade-off. 
This work provides a strong foundation for expanding sleep neurophysiology analysis using more accessible peripheral measurements.

\subsection*{AI use statement}

In this work, we used generative AI tools to provide feedback on research methodology or experiments, and implement methods.
We have not used generative AI tools for other tasks with required disclosure, and they are not applicable to this work.
Additionally, we used generative AI tools to create or edit software code, summarize or analyze existing literature and edit a research paper to improve readability. 
We have reviewed all AI-assisted work. For example, we manually checked the ideas, sentences, existing literature suggested by AI for its factual accuracy and potential plagiarism. 
LLM-generated code was also verified and tested for correctness by 2 authors. 
We take responsibility for the final content of this work, including text, claims or artifacts produced with the aid of generative AI.

\subsection*{Reproducibility statement}
To ensure the reproducibility of our work, we provide the model specification (Appendix~\ref{sec:model}) and implementation details (Appendix~\ref{sec:implementation}). 
The summary information on datasets (Appendix ~\ref{sec:datasets}) and preprocessing steps (Section ~\ref{sec:experiments}) are also provided.

\bibliography{main}
\bibliographystyle{iclr2027_conference}

\clearpage
\appendix
\section{Datasets} \label{sec:datasets}
% Dataset description

A brief overview of the datasets used in this study is provided below. Table~\ref{tab:training_data} summarizes the main statistics of the datasets used for pretraining, while Table~\ref{tab:downstream_data} presents the datasets used for downstream evaluation. Table~\ref{tab:channels} reports the channels and sampling rates available for each dataset.

\begin{table*}[b]
    \centering
    \caption{Statistics of training recordings (minimum duration: 256 min). Stage counts cover all annotated 30-s epochs in retained recordings, whereas effective training duration includes only complete 256-min input windows.}
    \label{tab:training_data}
    \resizebox{\textwidth}{!}{
    \small
    \renewcommand{\arraystretch}{1.5}
    \begin{tabular}{c c c c c c | c c c c c c | c c c c c}
        \toprule
        \multirow{2}{*}{\textbf{Dataset}} & \multirow{2}{*}{\textbf{Records}} & \multirow{2}{*}{\textbf{Subjects}} & \multirow{2}{*}{\textbf{Age (years)}} & \multirow{2}{*}{\textbf{BMI}} & \multirow{2}{*}{\begin{tabular}{@{}c@{}}\textbf{Sex \%} \\ \textbf{(Female/Male)}\end{tabular}} & \multicolumn{6}{c|}{\textbf{Stage count}} & \multicolumn{5}{c}{\textbf{Stage ratio (\%)}} \\
        & & & & & & W & N1 & N2 & N3 & REM & Total & W & N1 & N2 & N3 & REM \\
        \midrule
        PHY-Train & 990 & 990 & 55.2 $\pm$ 14.3 & N/A & 33 / 67 & 145,317 & 135,210 & 371,718 & 101,551 & 113,765 & 867,561 & 17 & 16 & 42 & 12 & 13 \\
        PHY-Test & 989 & 989 & 54.8 $\pm$ 14.3 & N/A & 37 / 63 & \multicolumn{6}{c|}{N/A} & \multicolumn{5}{c}{N/A} \\
        \midrule
        SHHS1 & 3,655 & 3,655 & 63.1 $\pm$ 11.5 & 28.1 $\pm$ 5.2 & 52 / 48 & 737,584 & 136,122 & 1,517,256 & 471,476 & 516,169 & 3,378,607 & 22 & 4 & 45 & 14 & 15 \\
        SHHS2 & 2,547 & 2,547 & 67.6 $\pm$ 10.3 & 28.3 $\pm$ 5.0 & 54 / 46 & 682,217 & 106,861 & 1,102,583 & 302,150 & 395,181 & 2,588,992 & 26 & 4 & 43 & 12 & 15 \\
        \midrule
        KISS & 5,675 & 5,675 & 44.6 $\pm$ 14.6 & 25.6 $\pm$ 4.2 & 21 / 79 & 946,226 & 640,532 & 1,516,640 & 584,211 & 648,113 & 4,335,722 & 22 & 15 & 35 & 13 & 15 \\
        KVSS & 800 & 800 & 51.4 $\pm$ 14.5 & 26.8 $\pm$ 4.4 & 24 / 76 & 134,500 & 122,688 & 261,618 & 46,394 & 95,539 & 660,739 & 20 & 19 & 40 & 7 & 14 \\
        \midrule
        MrOS1 & 2,758 & 2,758 & 76.4 $\pm$ 5.5 & 27.1 $\pm$ 3.8 & 0 / 100 & 943,949 & 129,086 & 1,234,915 & 221,090 & 381,366 & 2,910,406 & 32 & 4 & 43 & 8 & 13 \\
        MrOS2 & 991 & 991 & 81.0 $\pm$ 4.4 & 26.9 $\pm$ 3.8 & 0 / 100 & 381,597 & 80,258 & 425,045 & 44,896 & 129,853 & 1,061,649 & 36 & 8 & 40 & 4 & 12 \\
        \midrule
        MESA & 1631 & 1631 & 69.4 $\pm$ 9.1 & 28.6 $\pm$ 5.5 & 53 / 47 & 478,846 & 165,620 & 680,574 & 117,046 & 215,291 & 1,657,377 & 29 & 10 & 41 & 7 & 13 \\
        \bottomrule
    \end{tabular}
    }
\end{table*}

        % \midrule        
        % ISRUC-SG1 & 100 & 100 & 51.1 $\pm$ 15.9 & - & 44 / 56 & 20,965 & 11,507 & 28,260 & 17,468 & 11,912 & 90,112 & 23 & 13 & 31 & 19 & 13 \\        
        % ISRUC-SG2 & 16 & 8 & 46.9 $\pm$ 17.5 & - & 25 / 75 & 2,232 & 2,164 & 2,887 & 2,534 & 2,007 & 11,824 & 19 & 18 & 24 & 21 & 17 \\
        % ISRUC-SG3 & 10 & 10 & 39.6 $\pm$ 9.6 & - & 10 / 90 & 1,787 & 1,222 & 2,624 & 1,983 & 1,088 & 8,704 & 20 & 14 & 30 & 23 & 13 \\

\begin{table*}[t]
    \centering
    \caption{The dataset statistics used for validation and test.}
    \label{tab:downstream_data}
    \resizebox{\textwidth}{!}{
    \small
    \renewcommand{\arraystretch}{1.5}
    \begin{tabular}{c c c c c c | c c c c c c | c c c c c}
        \toprule
        \multirow{2}{*}{\textbf{Dataset}} & \multirow{2}{*}{\textbf{Records}} & \multirow{2}{*}{\textbf{Subjects}} & \multirow{2}{*}{\textbf{Age (years)}} & \multirow{2}{*}{\textbf{BMI}} & \multirow{2}{*}{\begin{tabular}{@{}c@{}}\textbf{Sex \%} \\ \textbf{(Female/Male)}\end{tabular}} & \multicolumn{6}{c|}{\textbf{Stage count}} & \multicolumn{5}{c}{\textbf{Stage ratio (\%)}} \\
        & & & & & & W & N1 & N2 & N3 & REM & Total & W & N1 & N2 & N3 & REM \\
        \midrule
        SHHS1-Val & 80 & 80 & 61.6 $\pm$ 12.0 & 28.0 $\pm$ 4.3 & 54 / 46 & 16,396 & 2,912 & 32,636 & 9,311 & 10,882 & 72,137 & 23 & 4 & 45 & 13 & 15 \\
        SHHS1-Test & 290 & 290 & 63.9 $\pm$ 11.1 & 28.4 $\pm$ 4.9 & 49 / 51 & 57,933 & 11,558 & 122,037 & 35,274 & 42,105 & 268,907 & 22 & 4 & 45 & 13 & 16 \\
        \midrule
        MESA-Val & 170 & 170 & 69.1 $\pm$ 8.6 & 28.9 $\pm$ 5.4 & 59 / 41 & 49,657 & 16,265 & 72,398 & 12,467 & 22,476 & 173,263 & 29 & 9 & 42 & 7 & 13 \\
        MESA-Test & 62 & 62 & 68.8 $\pm$ 8.9 & 29.6 $\pm$ 6.0 & 52 / 48 & 18,569 & 6,281 & 24,570 & 5,406 & 8,373 & 63,199 & 29 & 10 & 39 & 9 & 13 \\
        \midrule
        \edit{CFS-Train} & \edit{452} & \edit{452} & \edit{43.1} $\pm$ \edit{19.8} &
        \edit{32.0} $\pm$ \edit{9.3} & \edit{52 / 48} & \edit{101,447} & \edit{15,927} & \edit{175,733} & \edit{58,789} & \edit{59,978} & \edit{411,874} & \edit{25} & \edit{4} & \edit{43} & \edit{14} & \edit{15} \\
        \edit{CFS-Val} & \edit{50} & \edit{50} & \edit{40.4} $\pm$ \edit{19.0} &
        \edit{31.4} $\pm$ \edit{8.9} & \edit{54 / 46} & \edit{12,319} & \edit{1,751} & \edit{19,718} & \edit{5,879} & \edit{6,099} & \edit{45,766} & \edit{27} & \edit{4} & \edit{43} & \edit{13} & \edit{13} \\
        \edit{CFS-Test} & \edit{100} & \edit{100} & \edit{38.1} $\pm$ \edit{19.3} &
        \edit{31.6} $\pm$ \edit{9.0} & \edit{62 / 38} & \edit{21,456} & \edit{3,044} & \edit{37,974} & \edit{13,120} & \edit{11,997} & \edit{87,591} & \edit{25} & \edit{4} & \edit{43} & \edit{15} & \edit{13} \\
        
        % \edit{KISS-Val} & \edit{693} & \edit{693} & \edit{44.7} $\pm$ \edit{14.1} & \edit{26.0} $\pm$ \edit{4.3} & \edit{22 / 78} & \edit{121,319} & \edit{87,032} & \edit{187,347} & \edit{71,362} & \edit{78,128} & \edit{545,188} & \edit{22} & \edit{16} & \edit{35} & \edit{13} & \edit{14} \\
        \bottomrule
    \end{tabular}
    }
\end{table*}
\begin{table}[ht]
    \centering
    \caption{Channels and sampling rates included in different modalities across datasets. The manufacturer of PSG recording machine is also provided. When sampling rates are same across different channels in the modality, we only write once at the top row.}
    \label{tab:channels}
    \resizebox{\textwidth}{!}{%
    \begin{tabular}{lcccccccccccccc}
        \toprule
        \multirow{2}{*}{\textbf{Modality}}  
        & \multicolumn{2}{c}{\textbf{KISS / KVSS}} 
        & \multicolumn{2}{c}{\textbf{SHHS1}} 
        & \multicolumn{2}{c}{\textbf{SHHS2}} 
        & \multicolumn{2}{c}{\textbf{PHY}} 
        & \multicolumn{2}{c}{\textbf{MESA}} 
        & \multicolumn{2}{c}{\textbf{MrOS1 / 2}} 
        & \multicolumn{2}{c}{\edit{\textbf{CFS}}} \\ 
        
        \cmidrule(lr){2-3} 
        \cmidrule(lr){4-5} 
        \cmidrule(lr){6-7} 
        \cmidrule(lr){8-9} 
        \cmidrule(lr){10-11} 
        \cmidrule(lr){12-13}
        \cmidrule(lr){14-15}
        
         & Ch. & SR (Hz) 
         & Ch. & SR (Hz) 
         & Ch. & SR (Hz) 
         & Ch. & SR (Hz) 
         & Ch. & SR (Hz) 
         & Ch. & SR (Hz) 
         & \edit{Ch.} & \edit{SR (Hz)} \\ 
         
        \midrule

        \multirow{6}{*}{EEG}  
          & C3-M2 & 200 
          & C3-A2 & 125 
          & C3-A2 & 125 
          & C3-M2 & 200 
          & C4-M1 & 256 
          & C3-M2 & 256 
          & \edit{C3-A2} & \edit{128} \\
          
          & C4-M1 &  
          & C4-A1 &  
          & C4-A1 &  
          & C4-M1 &  
          & Oz-Cz &  
          & C4-M1 &  
          & \edit{C4-A1} & \\
          
          & O1-M2 &  
          &  &  
          &  &  
          & O1-M2 &  
          & Fz-Cz &  
          & O1-M2 &  
          &  & \\
          
          & O2-M1 &  
          &  &  
          &  &  
          & O2-M1 &  
          &  &  
          & O2-M1 &  
          &  & \\
          
          &  &  
          &  &  
          &  &  
          & F3-M2 &  
          &  &  
          &  &  
          &  & \\
          
          &  &  
          &  &  
          &  &  
          & F4-M1 &  
          &  &  
          &  &  
          &  & \\ 
          
        \midrule

        ECG  
          & ECG & 200 
          & ECG & 125 
          & ECG & 125 
          & ECG & 250 
          & ECG & 256 
          & ECG1-ECG2 & 512 
          & \edit{ECG1-ECG2} & \edit{256/512} \\ 
          
        \midrule

        \multirow{4}{*}{Respiratory} 
          & Flow & 200 
          & Flow & 10 
          & Flow & 10 
          & Flow & 200 
          & Flow & 32 
          & Flow & 64 
          & \edit{Nasal Pres} & \edit{64/128} \\
          
          & Thermistor &  
          &  &  
          & Thorax &  
          & Thorax &  
          & Thorax &  
          & Thermistor & 16 
          & \edit{Airflow} & \edit{32} \\
          
          & Thorax &  
          &  &  
          & Abdomen &  
          & Abdomen &  
          & Abdomen &  
          & Thorax &  
          & \edit{Thor Effort} & \edit{32} \\
          
          & Abdomen &  
          &  &  
          &  &  
          &  &  
          & Press &  
          & Abdomen &  
          & \edit{Abdo Effort} & \edit{32} \\ 
          
        \midrule

        PPG
          & -- & -- 
          & -- & -- 
          & -- & -- 
          & -- & -- 
          & Pleth & 256 
          & -- & -- 
          & \edit{PlethWV} & \edit{128} \\ 
        \midrule

        Manufacturer 
          & \multicolumn{2}{c}{Nox, Embla} 
          & \multicolumn{2}{c}{Compumedics} 
          & \multicolumn{2}{c}{Compumedics} 
          & \multicolumn{2}{c}{Unknown} 
          & \multicolumn{2}{c}{Compumedics} 
          & \multicolumn{2}{c}{Compumedics} 
          & \multicolumn{2}{c}{\edit{Compumedics}} \\ 
          
        \bottomrule
    \end{tabular}%
    }
\end{table}

\subsection{PhysioNet 2018 (PHY) Dataset}

The PhysioNet 2018 dataset was originally developed for the 2018 PhysioNet/CinC Challenge~\cite{goldberger2000physiobank, ghassemi2018you}. It was provided by the Computational Clinical Neurophysiology Laboratory and the Clinical Data Animation Laboratory at Massachusetts General Hospital. Although the main objective of the challenge was arousal detection, the dataset also contains expert-annotated sleep stages for 994 subjects. In addition, 991 recordings were reserved as a test set, although the corresponding labels remain private. In this study, both the labeled training set and the unlabeled test set were used for training \proposal. The original PhysioNet recordings were acquired at dataset-specific sampling rates, with most physiological signals sampled at 200 Hz. During preprocessing, the raw biosignals were resampled to 100 Hz to ensure a consistent sampling rate across datasets. Sleep stages were scored following the American Academy of Sleep Medicine (AASM) guidelines.

\subsection{Sleep Heart Health Study (SHHS) Dataset}

The Sleep Heart Health Study (SHHS)~\cite{zhang2018national, quan1997sleep} is a multicenter cohort study organized by the National Heart, Lung, and Blood Institute. The dataset contains data collected during two separate visits. SHHS1 corresponds to the first visit, conducted between 1995 and 1998, and included 6,441 men and women aged 40 and older. SHHS2 corresponds to the second visit, carried out between 2001 and 2003, and included 3,295 participants. Polysomnography (PSG) recordings were collected at home by trained technicians and included several physiological signals, such as EEG (C3-A2 and C4-A1), two EOG channels, EMG, respiratory effort, airflow, oxygen saturation, ECG, and body position.

For our experiments, the SHHS1 dataset was divided into 3,655 subjects for training, 80 subjects for validation, and 290 subjects for the final held-out test set. For survival prediction, nine training subjects were excluded due to unavailable features or zero follow-up time, and one test subject with zero follow-up time was excluded, yielding 3,646 training, 80 validation, and 289 test subjects for analysis. The test set was restricted to subjects belonging to the common evaluation set shared between our pipeline and the wav2Sleep baseline, allowing a consistent comparison between the two models. Subjects from the original held-out test set that were not part of this common subset were kept unused. The  SHHS2 dataset was used for pretraining and was not used for downstream tasks.

\subsection{Korea Image-based Sleep Study (KISS) Dataset}

The Korea Image-based Sleep Study (KISS) dataset~\cite{jeong2023standardized} is a standardized polysomnography (PSG) dataset collected between 2013 and 2020 across four sleep centers. Recordings were acquired using Embla and NOX-A1 PSG systems, resulting in a total of 10,253 recordings. Sleep stages were scored by experts according to the AASM version 2.6 guidelines~\cite{berry2012aasm, berry2017aasm}. Each recording contains 21 different biosignals, including several EEG, EOG, and EMG channels, as well as respiratory and movement-related signals. The dataset is publicly available through AI Hub~\cite{aihub}.

The training split defined in ~\cite{jeong2023standardized} are used for training. KISS was not used for downstream evaluation or final testing. 

\subsection{Korea Video Sleep Study (KVSS) Dataset}

The Korea Video Sleep Study (KVSS) dataset is a retrospectively constructed multicenter clinical cohort that includes synchronized infrared sleep video and polysomnography (PSG) recordings with expert annotations ~\cite{choi2026non}. Data were collected under IRB-approved protocols at three hospitals: Chungnam National University Hospital, The Catholic University of Korea St.~Vincent's Hospital, and Hallym University Hospital. The complete KVSS cohort contains 936 PSG examinations with synchronized infrared video, while 881 examinations were retained in this study after screening and quality control.
The recordings were obtained during routine clinical sleep studies for various sleep-related conditions, including suspected obstructive sleep apnea, insomnia, periodic limb movement disorder, and REM sleep behavior disorder. Infrared video was recorded simultaneously with PSG to maintain temporal synchronization. Although camera position and lighting conditions differed slightly between sites, all videos were standardized to MP4 format with a resolution of $640\times480$ pixels at 5~fps. PSG recordings were stored in European Data Format (EDF). Sleep stages (Wake, N1, N2, N3, and REM) and AASM-defined events were annotated by certified sleep technologists and reviewed by sleep physicians. The dataset was de-identified, and expert cross-checking was performed to verify synchronization and correct obvious scoring inconsistencies. In our experiments, KVSS was used as a PSG dataset for pretraining. Only the physiological signals required by the model were processed, while the synchronized infrared video data were not used.

\subsection{Multi-Ethnic Study of Atherosclerosis (MESA) Dataset}

The Multi-Ethnic Study of Atherosclerosis (MESA)~\cite{chen2015racial} is a multicenter cohort including 6,814 adults aged between 45 and 84 years from four racial and ethnic groups (White, Black, Hispanic, and Chinese-American participants). As part of Exam 5, conducted between 2010 and 2013, the MESA Sleep study enrolled 2,237 participants who completed a single night of unattended in-home polysomnography (PSG) together with wrist actigraphy. PSG recordings were set up during evening home visits by trained staff, while sleep stages and respiratory events were scored at a centralized sleep reading center using standardized procedures. The National Sleep Research Resource (NSRR) provides the PSG recordings in European Data Format (EDF), together with XML annotation files containing sleep-stage and respiratory-event information. Respiratory-event annotations were harmonized through rule-based post-processing of the original labels to ensure consistent criteria across datasets, following the procedure described in~\cite{ahn2025refining}. The PSG recordings include EEG (Fz-Cz, Cz-Oz, and C4-A1), EOG, EMG, ECG, nasal airflow, thoracic and abdominal respiratory effort, oxygen saturation, and body position. In our experiments, the original MESA training split was kept unchanged with 1,631 recordings. To ensure a consistent comparison with the wav2sleep baseline, the validation set was filtered to 170 recordings, while the final held-out test set was restricted to 62 recordings shared between the two evaluation pipelines. Recordings from the original test set that were not part of this common subset were kept unused.

\subsection{Osteoporotic Fractures in Men Study (MrOS) Dataset}

The Osteoporotic Fractures in Men Sleep Study (MrOS Sleep)~\cite{blackwell2011associations} is a multicenter cohort of older men aged 65 years and above. It includes unattended in-home polysomnography (PSG), which was set up by trained technicians and annotated using standardized scoring procedures. The study was conducted between December 2003 and March 2005, during which 3,135 participants from the original MrOS cohort of 5,994 men completed overnight unattended PSG. The dataset available through the National Sleep Research Resource (NSRR) provides PSG recordings in European Data Format (EDF), together with XML annotation files. These recordings contain multiple physiological signals, including EEG, EOG, EMG, ECG, nasal airflow, thoracic and abdominal respiratory effort, oxygen saturation, and body position. Respiratory-event annotations were harmonized using rule-based post-processing of the original labels to maintain consistent criteria across datasets, following the procedure described in~\cite{ahn2025refining}. In our experiments, data from both MrOS visits were included in the pretraining dataset. For MrOS1, the C3-A2 EEG channel and abdominal respiratory effort were used, resulting in 2,758 recordings. For MrOS2, the C3-M2 EEG channel and abdominal respiratory effort were used, resulting in 991 recordings. A subset of MrOS2 recordings exhibited a different signal amplitude scale and was harmonized during preprocessing.

\subsection{Cleveland Family Study (CFS) Dataset}

The Cleveland Family Study (CFS)~\citep{zhang2018national, redline1995familial} is a family-based cohort designed to investigate familial risk factors associated with sleep apnea. The cohort includes 2,284 participants from 361 families who were followed across several visits over a period of approximately 16 years. The National Sleep Research Resource (NSRR) provides overnight polysomnography recordings, including full-night PSG data collected during Visit 5. Subjects younger than 14 years were excluded because pediatric sleep studies follow different clinical scoring criteria. The available PSG signals and corresponding sleep-stage annotations were processed using the same preprocessing pipeline applied to the other datasets. 

CFS is excluded from \proposal's training stages. For downstream evaluation, however, we train a separate task-specific head using labels from the CFS training split and evaluate it on the held-out CFS test split. Thus, the CFS results assess the transfer of a fixed translation model to an unseen cohort under cohort-specific supervised downstream evaluation, rather than zero-shot transfer of the complete prediction pipeline. In contrast, CFS is an internal dataset for the wav2sleep baseline, as it is included among the datasets used for model training and evaluation. Therefore, while the wav2sleep results are computed on its held-out CFS test split, CFS does not represent an unseen dataset for wav2sleep, unlike in our evaluation setting.

% \subsection{\edit{Study of Osteoporotic Fractures (SOF) Dataset}}

% \edit{The Study of Osteoporotic Fractures (SOF) \citep{zhang2018national, spira2008sleep} is a prospective, multi-center cohort originally established to study risk factors for osteoporotic fractures and falls in older women. The sleep study includes women aged 65--89 years and provides raw EDF PSG recordings together with PSG-derived variables through the NSRR. Sleep studies were completed on 461 SOF participants at Visit 8, and recordings passing our quality-control criteria were used for downstream evaluation. In this work, SOF is used exclusively as an external held-out downstream evaluation dataset and is not included in the pretraining corpus. We use the available PSG recordings and annotations for sleep stage classification and sleep-disordered breathing segmentation, enabling evaluation of cross-cohort generalization to an older, all-female population.}

\section{Model Specification} \label{sec:model}
Table~\ref{tab:model_specs} and Table~\ref{tab:cnnspec_specs} summarize \proposal and CNN-Transformer model (C-T baseline) specifications, respectively.
C-T baseline is trained with the same training recipe as translation model, described in Table~\ref{tab:hyper_training}.

We also evaluated larger C-T baseline variants with increasing model capacity and window size. 
These variants did not converge reliably under the varied training configurations. 
% We therefore report the smaller C-T baseline and provide the larger-model configurations. 
One of the tested larger model specification and training results are provided in Table~\ref{tab:cnnspec_specs_larger}.
This observation is limited to the settings tested and does not rule out improvements from further optimization.

\section{Implementation Details} \label{sec:implementation}
We provide training hyperparameters in Table~\ref{tab:hyper_training}. The hyperparameters used for each downstream tasks are summarized in Table~\ref{tab:hyper_downstream}. We have trained and evaluated \proposal and baseline methods on a single NVIDIA H200 GPU (140 GiB HBM3e) with two Intel Xeon Platinum 8462Y+ CPUs. Models are implemented in PyTorch 2.5.1 (CUDA 12.1, cuDNN 9.1) with PyTorch Lightning 2.6.0, and trained in bf16 mixed precision with distributed data parallelism. End-to-end training takes approximately 10 hours per single modality.
% Chunking: 20min/epoch (Source + Target)
% Wordization: 20min/epoch (Source + Target)
% Translation: 40min/epoch (Together)

\textbf{PPG pretraining}. PPG channel is not included in SleepMaMi configuration and is only available in MESA dataset. 
To tackle this, we adopt two-stage pretraining strategy.
First, we pretrain SleepMaMi using three modalities (EEG, ECG and Respiratory) on the full training dataset.
We then introduce a PPG-specific private encoder and fine-tune only this encoder on the MESA PPG data while keeping all other modules frozen.
This procedure maps PPG inputs into the common representation space learned from the other modalities, enabling alignment across all four modalities.

\textbf{Sleep-stage classification.} We use a CNN-TCN architecture. CNN layers extract local spectral patterns from the reconstructed EEG spectrogram. The TCN then models temporal dependencies across consecutive sleep epochs using dilated convolutions. The final features are projected to four sleep-stage classes. The model is trained for 20 epochs with weighted cross-entropy loss.

\textbf{Wordbook initialization}.
As described in Section~\ref{sec:method}, the wordbook is initialized using data-driven approach.
Wordization is disabled for the first $I$ batches, during which chunk embeddings and their lengths are collected.
The dominant chunk lengths are then identified, and chunks are allocated to these length groups in proportion to the number of chunks.
Within each group, the collected chunk embeddings are clustered once using Lloyd's algorithm to initialize the corresponding \emph{words}.

$\alpha$ is calibrated as the median distance between each chunk to the nearest centroid of a different length group, divided by the smallest log-duration gap between initial length groups.
A smaller value may underweight duration, causing chunks with similar contents but different lengths to be assigned to the same word.
In contrast, a large $\alpha$ may cause the duration to dominate content, making assignment primarily length-dependent. 
This calibration balances the contributions of content and duration to word assignment.

\textbf{PasL reproduction}. We reproduce PasL using the architecture and training setup described in the original paper. EEG spectrograms have a resolution of $256\times512$. A VQGAN tokenizer maps each spectrogram to an $8\times64$ grid of 512 discrete tokens using an 8,192-entry codebook. The masked Transformer uses an 8-layer encoder and decoder, 768-dimensional hidden states, and 8 attention heads. The model is trained for 500 epochs with AdamW using the masked-token cross-entropy (CE) objective. Reconstructed EEG spectrograms are obtained with the frozen VQGAN decoder. Training converges stably, although our reproduced downstream performance remains slightly below the values reported in the original paper.

% is calibrated so that the separation along the length channel is at least as large as the median content distance.  
% This calibration encourages chunks with similar content but different lengths are encouraged to be assigned different words. 
% The wordbook is learned for each modality to maintain the unique features of each modality.

\begin{table}[t]
    \centering
    \caption{\proposal model specifications. $D$ denotes the shared latent width; one base token covers 1\,s of signal (30 tokens per epoch), $L'$ is the number of base tokens and $C^m$ is the number of words in a full night recording.}
    \label{tab:model_specs}
    \resizebox{0.99\linewidth}{!}{%
    % \small
    \begin{tabular}{@{}lllll@{}}
    \toprule
    \textbf{Component} & \textbf{Module} & \textbf{Configuration} & \textbf{Output} & \textbf{Notes} \\    
    \midrule
    Frozen Base Encoder & SleepMaMi & Micro-encoder private and shared modules & $(L', 256)$ \\    
    \midrule
    \quad Parameters & \multicolumn{4}{l}{$\approx$20.4M total } \\
    \midrule
    Chunking & Chunk encoder & $2\times$ Mamba blocks & $(L', 256)$ & \\
    & & ($d_{\mathrm{state}}{=}16$, conv=4,  expand=2, Gated MLP with $\times 4$ expansion) & & \\
     & Router &  $W_q, W_k \in \mathbb{R}^{256\times256}$ & $(L', 256)$ \\
    \midrule
    Encoder & Encoder & $4\times$(2 sliding-window Transformer + 1 Full Transformer) & $(C^m, 256)$ & \\
     & & $D{=}256$, 8 heads, Gated MLP with $\times 4$ expansion, RoPE at word start & & \\     
    \midrule
    Decoder & Block AR & Depth 2, Self + Cross attention per sync block & $(C^{EEG},|\mathcal{W}|+1)$   & \\
     & & id head, length head & $(C^{EEG},2r+1)$ &$r$: Residual slot size \\          
    \midrule
    Dechunking & Wordbook lookup & Scatter to the base grid (STE) & $(T, 256)$ & \\
     & Dechunk encoder & $2\times$ Mamba blocks & $(T, 256)$ & \\
    \midrule
    Reconstruction head  & Spectrogram head  & Conv1d($(k,s){=}(30,6)$) $\to$ Conv1d($(k,s){=}(5,5)$) & ($T_y,N_f$) & Only for EEG\\
                        &                   & \quad $\to$ Linear ($256{\to}N_f$) & & \\
                        & Upsampling head  & Conv1d($(k,s){=}(30,1)$) $\to$ Conv1d($(k,s){=}(30,1)$) & $(L,1)$ & for peripherals\\
                        &                   & \quad $\to$ Linear ($256{\to}f_s$) & $(L=L'\times N_f)$& \\
    \midrule
    \quad Parameters & \multicolumn{4}{l}{$\approx$25.8M total } \\
    \bottomrule
    \end{tabular}
    }
\end{table}

\begin{table}[t]
\centering
\caption{CNN-Transformer baseline specifications. $L$ is the total length of full-night vector (100 Hz $\times$ the number of seconds in full-night recording)}
\label{tab:cnnspec_specs}
    \resizebox{0.90\linewidth}{!}{%
    \small
    \begin{tabular}{@{}lll@{} }
    \toprule
    \textbf{Component} & \textbf{Configuration} & \textbf{Output} \\
    \midrule
    InstanceNorm & RevIN per 30\,s epoch, no affine & $(L, 1)$ \\
    Conv tokenizer & Conv1d $D: 1{\to}32{\to}64{\to}128{\to}256$, Depth: 1, 1, 4, 4 & $(L/250, 256)$ \\     
     & Kernel size: 10, 5, 1, 5  &  \\     
     & Stride is equal to kernel size for the last layer. Otherwise 1 is used.& \\
    % Refine module & ConvNext Layer $\times$ 4 & $(L/250, 256)$ \\     
    Reshape & Reshape seq from $(L/250, 256){\to}(240, 256)$ (Each seq is 20 epochs) & $(240, 256)$ \\
            & Batch size increases by $L/(240\times250)$ & \\
    Encoder & (2 sliding-window Transformer + 1 Full Transformer)$\times4$ & $(240, 256)$ \\
    & $D{=}256$, 8 heads, Gated MLP with $\times 4$ expansion, RoPE & \\
    Reshape & Return to the original shape & $(L/250, 256)$ \\
    Spectrogram head & Conv1d($(k,s){=}(12,3)$) $\to$ Conv1d($(k,s){=}(4,4)$)\quad $\to$ Linear ($256{\to}N_f$) & $T_y\times\,N_f$ \\
    % Training & spectrogram reconstruction only (no masking, no codebook) & \\
    \midrule
    \quad Parameters & \multicolumn{2}{l}{$\approx$14.9M total} \\
    \bottomrule
    \end{tabular}
    }
\end{table}

\begin{table}[t]
\centering
\caption{Larger CNN-Transformer baseline specifications and training results.}
\label{tab:cnnspec_specs_larger}
    \resizebox{0.90\linewidth}{!}{%
    \small
    \begin{tabular}{@{}lll@{} }
    \toprule
    \textbf{Component} & \textbf{Configuration} & \textbf{Output} \\
    \midrule
    InstanceNorm & RevIN per 30\,s epoch, no affine & $(L, 1)$ \\
    Conv tokenizer & Conv1d $D: 1{\to}32{\to}64{\to}128{\to}256$, Depth: 4, 8, 16 & $(L/250, 256)$ \\     
     & Kernel size: 10, 5, 5  &  \\     
     & Stride is equal to kernel size for the last layer. Otherwise 1 is used.& \\
    Refine module & ConvNext Layer $\times$ 6 & $(L/250, 256)$ \\     
    % Reshape & Reshape seq from $(L/250, 256){\to}(120, 256)$ (Each seq is 20 epochs) & $(120, 256)$ \\
            % & Batch size increases by $L/(120\times250)$ & \\
    Encoder & (2 sliding-window Transformer + 1 Full Transformer)$\times4$ & $(L/250, 256)$ \\
    & $D{=}256$, 8 heads, Gated MLP with $\times 4$ expansion, RoPE & \\
    % Reshape & Return to the original shape & $(L/250, 256)$ \\
    Spectrogram head & Conv1d($(k,s){=}(12,3)$) $\to$ Conv1d($(k,s){=}(4,4)$)\quad $\to$ Linear ($256{\to}N_f$) & $T_y\times\,N_f$ \\
    % Training & spectrogram reconstruction only (no masking, no codebook) & \\
    \midrule
    \quad Parameters & \multicolumn{2}{l}{$\approx$22.8M total} \\
    \midrule
    Training results & \multicolumn{2}{l}{Trained for 10 epochs, MAE: 0.0837, SNR: 11.8015, Test MF1: 0.170 on SHHS1} \\
    \bottomrule
    \end{tabular}
    }
\end{table}

\begin{table}[ht]
    \centering
    \caption{Hyperparameter settings for training.}
    \label{tab:hyper_training}
    \resizebox{0.95\linewidth}{!}{%
    \begin{tabular}{@{}llll@{}}
        \toprule
        \textbf{Stage} & \textbf{Item} & \textbf{Value} & \textbf{Notes} \\ \midrule
        \multirow{14}{*}{Context-aware chunking} & Batch size & 32 & PSG cases  \\         
         & Optimizer & AdamW &  \\
         & $\beta_1$ & 0.9 &  \\
         & $\beta_2$ & 0.99 &  \\
         & Weight decay ($\lambda$) & 0.01 &  \\
         & Initial learning rate & $2.00 \times 10^{-3}$ &  \\
         & Learning rate schedule & Cosine annealing with warmup &  \\
         & Final learning rate & $1.00 \times 10^{-8}$ &  \\
         & Training epochs & 5 &  \\
         & Patch size & 1 Second & \\
         & Target chunking length  & 4 &  \\         
         % & Weight for reconstruction loss ($\lambda_{recon}$) & 1.0 & \\
         & Weight for balancing ratio loss ($\lambda_{ratio}$) & 0.03 & \\          
         & Weight for Align loss ($\lambda_{align}^{m}$) & 0.2 & $m \in \{\text{ECG, PPG, Respiratory}\}$, 0 for EEG \\ 
         & Weight for barrier loss ($\lambda_{barrier}$) & 0.01 & \\          
         \midrule

         \multirow{10}{*}{Wordization} & Batch size & 32 & PSG cases  \\         
         & Optimizer & AdamW &  \\
         & $\beta_1$ & 0.9 &  \\
         & $\beta_2$ & 0.99 &  \\
         & Weight decay ($\lambda$) & 0.01 &  \\
         & Initial learning rate & $1.00 \times 10^{-4}$ &  \\
         & Learning rate schedule & Cosine annealing with warmup &  \\
         & Final learning rate & $1.00 \times 10^{-8}$ &  \\
         & Training epochs & 5 &  \\         
         % & Weight for reconstruction loss ($\lambda_{recon}$) & 1.0 & \\
         & Weight for quantization loss ($\lambda_{Q}$) & 0.5 & \\                  
         & Wordbook size ($K$) & 2048 & \\                  
         \midrule
        
        \multirow{10}{*}{Translation} & Batch size & 8 & PSG cases  \\         
         & Optimizer & AdamW &  \\
         & $\beta_1$ & 0.9 &  \\
         & $\beta_2$ & 0.99 &  \\
         & Weight decay ($\lambda$) & 0.01 &  \\
         & Initial learning rate & $2.00 \times 10^{-3}$ &  \\
         & Learning rate schedule & Cosine annealing with warmup &  \\
         & Final learning rate & $1.00 \times 10^{-8}$ &  \\
         & Training epochs & 5 &  \\
         & Block size for AR & 3 & Epochs (90 seconds)  \\
         & Weight for duration prediction loss ($\lambda_{dur}$) & 1.0 & \\          
         \bottomrule
    \end{tabular}
    }
\end{table}

\begin{table}[!t]
    \centering
    \caption{Training details used for downstream tasks.}
    \label{tab:hyper_downstream}
    \resizebox{0.90\linewidth}{!}{%
    \begin{tabular}{@{}llll@{}}
    \toprule
    \textbf{Downstream task} & \textbf{Item} & \textbf{Value} & \textbf{Notes} \\
    \midrule

    \multirow{6}{*}{Sleep staging}
        & Model & CNN-TCN & PSG cases\\
        & Loss & Weighted Cross Entropy & \\
        & Optimizer & AdamW & \\
        & Initial learning rate & $3.00 \times 10^{-4}$ & \\
        & Weight decay & $1.00 \times 10^{-2}$ & \\
        & LR schedule & Cosine annealing & \\
    \midrule

    \multirow{5}{*}{Arousal detection}
        & Model & Linear ($256 \rightarrow 2$) & PSG cases \\
        & Loss & CrossEntropy  \\
        & Optimizer & Adam & \\
        & Initial learning rate & $1.00 \times 10^{-2}$ & \\
        & Weight decay & $1.00 \times 10^{-4}$ & \\
    \midrule

    \multirow{6}{*}{Survival prediction}
        & Model & Linear ($256 \rightarrow 3$) & PSG cases \\
        & Input & Full-night mean pooling  \\
        & Loss & Cox PH & \\
        & Optimizer & AdamW & \\
        & Initial learning rate & $1.00 \times 10^{-2}$ & \\
        & Weight decay & $1.00 \times 10^{-4}$ & \\
    \midrule

    \multirow{2}{*}{Common settings}
        & Batch size & 8  \\
        & Training epochs & 20 & \\

    \bottomrule
    \end{tabular}%
    }
\end{table}

\section{EEG Spectrogram} \label{sec:spectrogram}
In \proposal, we model the target EEG as a time-frequency representation rather than as the raw waveform. 
Although peripheral signals encode information coupled to brain activity, they are unlikely to support reliable recovery of the fine-grained, sample-level details of EEG waveforms. 
In contrast, the EEG spectrogram provides a physiologically meaningful and temporally localized representation of spectral activity that is more suitable for cross-modal estimation, as suggested by \citet{zha2026physiology}.

The EEG spectrogram is built in the following procedure. 
Each EEG recording is normalized by the standard deviation of each dataset's training split and partitioned into non-overlapping windows of $\omega$ seconds, from which log-power multi-taper spectra~\citep{prerau2017sleep} are computed over 0.5-32 Hz. 
Any windows carrying no usable signal (flat or non-finite, e.g., a detached electrode) are set to zero. 
The dB values are subsequently mapped to [0,1] using the 0.1st and 99.9th percentiles measured over the training corpus.
The same percentile-based mapping is applied across datasets and splits after dataset-specific amplitude normalization. 
Accordingly, the resulting values represent power on a common normalized scale.
% The same range is applied to the validation and test splits, so a given value corresponds to the same physical power in every recording. 
We use $\omega$ of five seconds as our default setting, except the reconstruction experiment where thirty seconds is used for fair comparison to other baselines.

\section{Modifications to SleepMaMi architecture} \label{sec:sleepmami}
We use the shared embeddings from SleepMaMi~\citep{park2026sleepmami}, with several architectural modifications to support translation from a single source modality.
In the original architecture, the shared module receives the concatenated embeddings of all available modalities.
Consequently, generating shared embeddings requires access to multiple modalities at once, including the target modality - this is not permissible in our translation setting.
We therefore modify the shared module to process each modality independently and align the resulting representations across modalities.
This design allows the model to encode a source signal without requiring the target signal while preserving a common cross-modal representation space.

We also modify the original leave-one-out contrastive learning (LOO-CL) objective. 
For each source modality, we randomly sample a subset of the remaining modalities and contrast the representations averaged over them with that of the source modality.
Exposing the model to different modality subsets encourages it to capture physiological features that are consistent across diverse combinations of modalities.

Finally, we modify the fusion module to combine the shared embeddings obtained from a randomly selected subset of modalities with the private embeddings.
This modification further encourages the shared representations to encode modality-agnostic information.

\begin{figure}[h!]
    \centering

    \begin{subfigure}[t]{0.47\linewidth}
        \centering
        \includegraphics[width=\linewidth,bb=0 0 2459 1294]{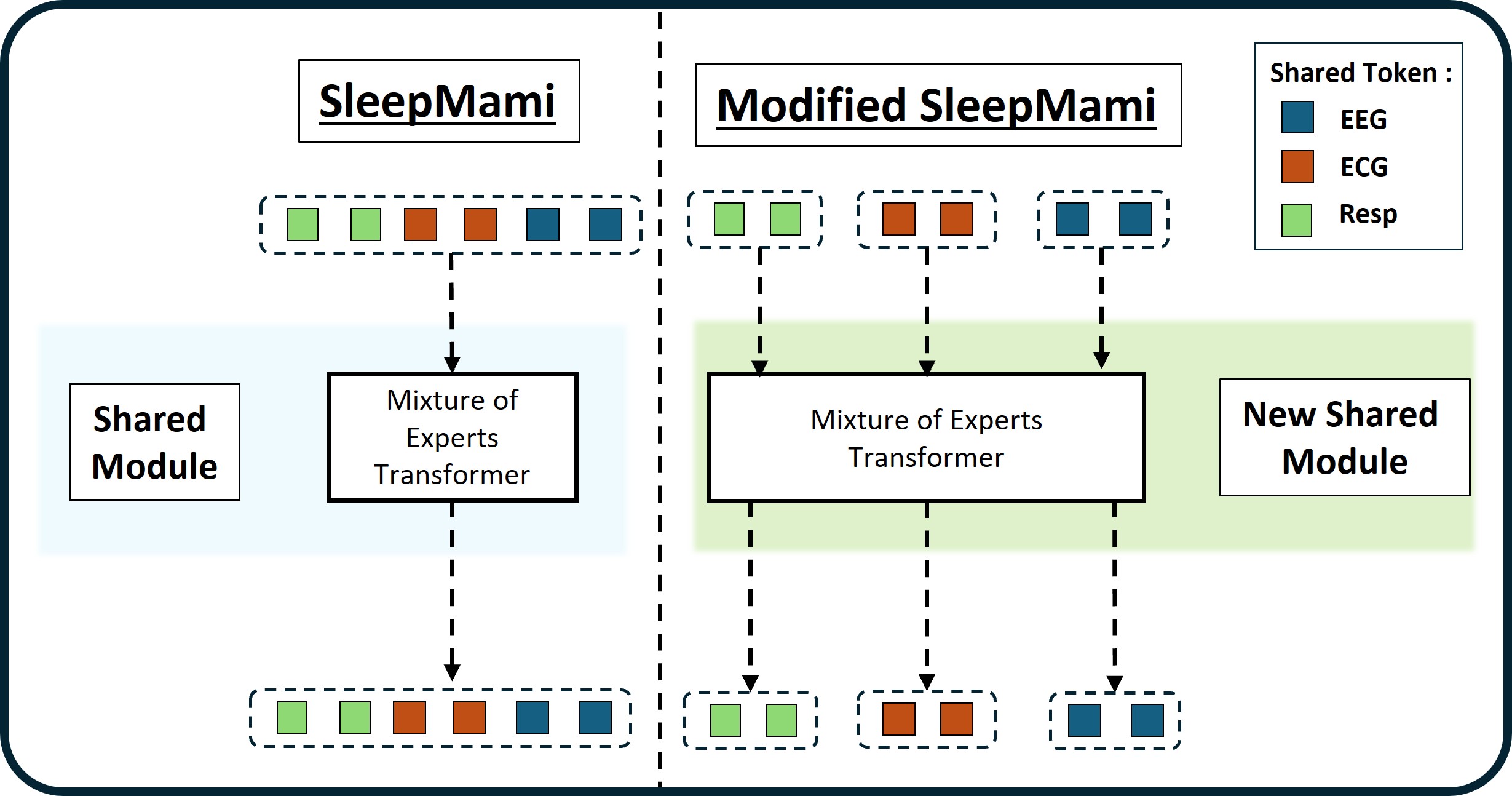}
        \caption{Shared module modification}
        \label{fig:shared_module_modification}
    \end{subfigure}
    \hfill
    \begin{subfigure}[t]{0.47\linewidth}
        \centering
        \includegraphics[width=\linewidth,bb=0 0 2368 1252]{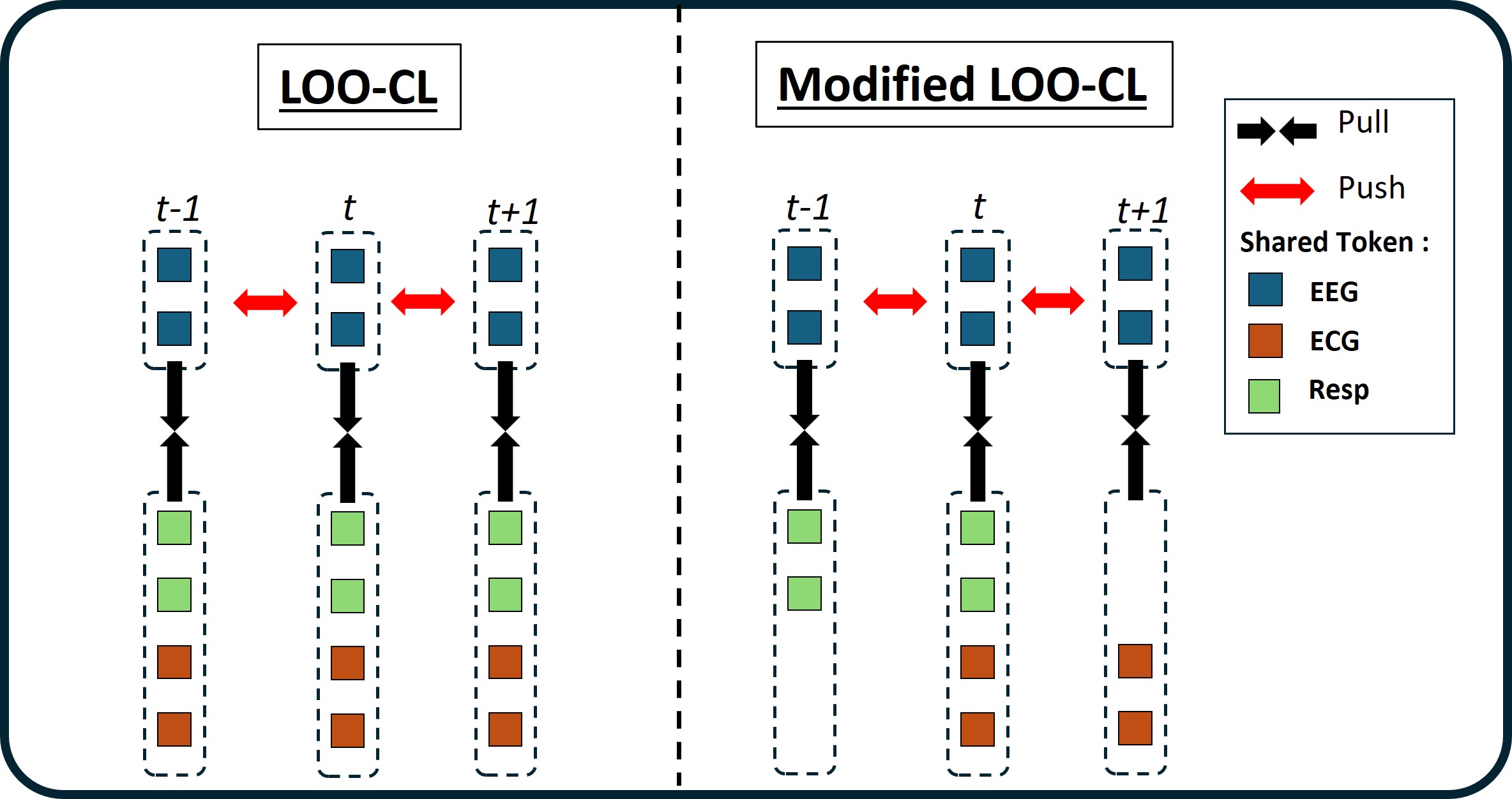}
        \caption{LOO-CL objective modification}
        \label{fig:loo_cl_modification}
    \end{subfigure}

    \caption{
    Overview of the modifications to the SleepMaMi architecture.
    (a) The shared module processes each modality independently, enabling
    single-modality inputs while preserving a common representation space.
    (b) LOO-CL is modified to handle different subsets of available modalities.
    }
    \label{fig:sleepmami_modifications}
\end{figure}

\section{Auxiliary losses} \label{sec:losses}

This section provides a more detailed description of auxiliary losses used during the context-aware chunking stage.

\textbf{Ratio loss ($\mathcal{L}_{\mathrm{ratio}}$)}: We adopt this loss to regulate the proportion of predicted boundaries. The formulation follows \citet{hwang2026dynamic}, where $N$ controls the target chunking ratio.
\begin{equation}
    \begin{aligned}
         \mathcal{L}_{ratio} &= \frac{N}{N-1} ((N-1)FG + (1-F)(1-G)), \quad F=\frac{1}{L'}\sum_i \mathbbm{1}_{p_i\ge0.5}, \quad G=\frac{1}{L'}\sum_i p_i \\
         % \mathcal{L}_{align} &= \sum_{(s, l_s) \in \mathcal{S}} \|Avg(\mathbf{z}_{s:s+l_s}) - Avg(\mathbf{e}_{s:s+l_s})\|^2 \\
         % \mathcal{L}_{barrier} &= \sum_{j=1}^{L'/B} \text{log}(p_{B \cdot j})         
     \end{aligned}
\end{equation}

\textbf{Alignment loss ($\mathcal{L}_{\mathrm{align}}$)}: This loss prevents the refined embeddings of chunk encoder from deviating substantially to modality-specific representation space. 
\begin{equation}
    \begin{aligned}         
         \mathcal{L}_{align} &= \sum_{c=1}^{C} \|\bar{\mathbf{z}}_c - \bar{\mathbf{e}}_c\|_2^2, \quad \bar{\mathbf{e}}_{c}=\frac{1}{l_c}\sum_{t=s_c}^{s_c+l_c-1}\mathbf{e}_t \\         
     \end{aligned}
\end{equation}

\textbf{Barrier loss ($\mathcal{L}_{\mathrm{barrier}}$)}: This loss encourages boundaries to be inserted at regular temporal intervals, thereby the resulting chunks are well placed for the block autoregressive decoder.
\begin{equation}    
      \mathcal{L}_{\mathrm{barrier}} = -\sum_{a \in \mathcal{A}} \text{log}(p_{a}), \quad \text{where} \quad \mathcal{A} = \{1+jB: j \ge 1, 1+jB \le L'\}     
\end{equation}

%%%%%%%%%%%%%%%%%%%%%%%%%%%%%%%%%%%%%%%%%%%%%%%%%%%%%%%%%%%%%%%%%%%%%%%%%%%%%%%%%%%%%%%%%%%%%%%
\section{Ablation} \label{sec:ablation}
We conduct ablation study to investigate the individual contribution of our core components.
% We examine three key hyperparameters: the wordbook size ($K$), target chunking ratio ($N$) and block size ($B$ (in epochs)).

\textbf{Loss components}. 
Table~\ref{tab:ablation_loss} provides the ablation results for the auxiliary loss components.
We conduct this experiment using the SHHS1 training set and evaluate respiratory-to-EEG sleep staging accuracy on SHHS1 under the linear-probing protocol. 
Removing $\mathcal{L}_{\mathrm{Barrier}}$ results in accuracy decrease of 0.007 (0.7 percentage points), demonstrating its contribution to maintaining temporal alignment between the source and target sequences.
Removing $\mathcal{L}_{\mathrm{Align}}$ and $\mathcal{L}_{\mathrm{Barrier}}$ decreases accuracy by 0.104 (10.4 percentage points), suggesting that the alignment objective prevents the refined chunk representations from drifting toward modality-specific features.
We do not report the configuration without  $\mathcal{L}_{\mathrm{Ratio}}$, since removing it causes the chunking mechanism to collapse and does not provide meaningful results.

\begin{table}[t]
\centering
\caption{Ablation on loss components.}
\label{tab:ablation_loss}
\begin{tabular}{cccccc}
\toprule
\textbf{Index} &
\textbf{$\mathcal{L}_{\mathrm{Recon}}$} &
\textbf{$\mathcal{L}_{\mathrm{Ratio}}$} &
\textbf{$\mathcal{L}_{\mathrm{Align}}$} &
\textbf{$\mathcal{L}_{\mathrm{Barrier}}$} &
\textbf{Acc.} \\
\midrule
1 & \checkmark & \checkmark &            &            & 0.649 \\
2 & \checkmark & \checkmark & \checkmark &            & 0.746 \\
3 & \checkmark & \checkmark & \checkmark & \checkmark & \textbf{0.753} \\
\bottomrule
\end{tabular}
\end{table}

\textbf{Chunking mechanism}.
To assess the effectiveness of \emph{context-aware chunking} and \emph{wordization}, we compare \proposal against two alternative methods:
\emph{Base2base} performs no chunking or wordization and directly translates the source base embeddings $\mathbf{e}^m$ into the target base embeddings $\mathbf{e}^{EEG}$.
\emph{FixedPool} replaces context-aware chunking with average pooling over non-overlapping windows of four base embeddings, matching the chunking ratio of $N=4$.
Table~\ref{tab:ablation_chunking} provides respiratory-to-EEG sleep staging accuracy on SHHS1 under the linear-probing protocol, using SHHS1 training set. 
\proposal outperforms Base2base by 0.022 (2.2 percentage points), demonstrating that word-level abstraction improves downstream performance.
It also surpasses FixedPool by 0.020 (2.0 percentage points), showing that adaptively locating chunk boundaries preserves more task-relevant temporal structure at a comparable compression level.
These gains highlight that the benefit of \proposal arises not only from shorter sequences and improved computational efficiency, but also from learning context-aware, physiologically meaningful abstractions.

\textbf{Duration awareness and duration residual prediction}. 
Each word in \proposal consists of a content component $\nu_k\in\mathbb{R}^d$ and a duration component $\eta_k \in \mathbb{R}$.
The decoder additionally predicts an occurrence-specific duration residual to account for variations among chunks assigned to the same word.
Table~\ref{tab:ablation_duration} evaluates the contributions of these two components.
We conduct this experiment using the SHHS1 training set and report respiratory-to-EEG sleep staging accuracy on SHHS1 under the linear-probing protocol. 

In the content-only variant, the word identity is determined solely by the content.
Incorporating the duration term $\eta_k$ in the wordbook enhances accuracy from 0.732 to 0.748, an increase of 0.016 (1.6 percentage points).
This result suggests that explicitly distinguishing chunks with similar content but different durations improves temporal alignment during translation.
Residual prediction further improves accuracy by 0.005 (0.5 percentage points), indicating that modeling deviations from the prototypical duration helps reduce residual temporal discrepancy.

\begin{table}[t]
\centering
\begin{minipage}[t]{0.45\linewidth}
    \centering
    \small
    \caption{Ablation on chunking mechanism.}
    \label{tab:ablation_chunking}
    \begin{tabular}{lc}
    \toprule
    \textbf{Item} & \textbf{Acc.} \\
    \midrule
    Base2base     & 0.731 \\
    FixedPool & 0.733 \\
    \proposal    & \textbf{0.753} \\
    \bottomrule
    \end{tabular}
\end{minipage}
\begin{minipage}[t]{0.5\linewidth}
    \centering
    \caption{Ablation on duration modeling.}
    \label{tab:ablation_duration}
    \begin{tabular}{lc}
        \toprule
        \textbf{Item} & \textbf{Acc.} \\
        \midrule
        Content-only wordbook                           & 0.732 \\
        + Content- and duration-aware wordbook          & 0.748 \\
        + Occurrence-specific residual prediction       & \textbf{0.753} \\
        \bottomrule
    \end{tabular}
\end{minipage}
\end{table}

\textbf{Wordbook size}. Table~\ref{tab:ablation_K} provides the effect of varying $K$. 
We conduct this experiment using the full training set and evaluate respiratory-to-EEG sleep staging accuracy on SHHS1 under the linear-probing protocol, while fixing $B=3$ and $N=4$.
Performance improves as the wordbook size increases, with the gains saturating at $K=2048$. 
This suggests that a sufficiently large vocabulary is required to capture diverse patterns, while further increasing its size provides limited benefit.

\textbf{Target chunking ratio}. Table~\ref{tab:ablation_N} provides the results for different values of $N$. 
We conduct this experiment using the full training set and evaluate respiratory-to-EEG sleep staging accuracy on SHHS1 under the linear-probing protocol, while fixing $K=2048$ and $B=3$.
Increasing the chunking ratio up to $N=4$ does not harm the model performance, but larger values lead to a modest decrease in accuracy.
These results suggest that the wordization mechanism can effectively compress the target sequence while preserving task-relevant information.
The experiment is done under linear probing protocol using the full training set and respiratory-to-EEG sleep staging accuracy on SHHS is reported.  
% We fix $K=2048$ and $B=3$ in this experiment.

\textbf{Block size}. 
Table~\ref{tab:ablation_B} provides the effect of varying $B$. 
We conduct this experiment using the full training set and evaluate respiratory-to-EEG sleep staging accuracy on SHHS1 under the linear-probing protocol, while fixing $K=2048$ and $N=4$.
Accuracy decreases as the block size increases, potentially because temporal misalignment accumulates over longer generation spans. 
This result highlights the importance of introducing block boundaries at appropriate intervals to limit temporal drift.

\begin{table}[t]
\centering
\begin{minipage}[t]{0.29\linewidth}
\centering
\small
\caption{Ablation on $K$.}
\label{tab:ablation_K}
\resizebox{0.6\linewidth}{!}{%
    \begin{tabular}{cc}
    \toprule
    $K$ & \textbf{Acc.} \\
    \midrule
    256  & 0.710 \\
    512  & 0.746 \\
    1024 & 0.752 \\
    \textbf{2048} & \textbf{0.757} \\
    \bottomrule
    \end{tabular}
}
\end{minipage}
\hfill
\begin{minipage}[t]{0.29\linewidth}
\centering
\small
\caption{Ablation on $N$.}
\label{tab:ablation_N}
\resizebox{0.5\linewidth}{!}{%
    \begin{tabular}{cc}
    \toprule
    $N$ & \textbf{Acc.} \\
    \midrule
    \textbf{2}  & \textbf{0.757} \\
    \textbf{4}  & \textbf{0.757} \\
    8  & 0.740 \\
    16 & 0.754 \\
    \bottomrule
    \end{tabular}
}
\end{minipage}
\hfill
\begin{minipage}[t]{0.4\linewidth}
\centering
\small
\caption{Ablation on $B$}
\label{tab:ablation_B}
\resizebox{0.5\linewidth}{!}{%
    \begin{tabular}{cc}
    \toprule
    $B$ (epochs) &  \textbf{Acc.} \\
    \midrule
    \textbf{3}  & \textbf{0.757} \\
    5  & 0.746 \\
    10 & 0.729 \\
    \bottomrule
    \end{tabular}
}
\end{minipage}

\end{table}

\section{Compression Ratio} \label{sec:compression}
The modality-specific wordbooks reduce sequence length by representing each variable-length chunk with a single token.
We quantify this effect using pretrained checkpoints and 50 subjects from SHHS1 or MESA (PPG).
Specifically, we define the effective compression ratio as $100(1-C/L')$, where $L'$ and $C$ denote the number of base patches and resulting word tokens, respectively.
All measurements are obtained with $N=4$.

As presented in Table~\ref{tab:compression}, the wordbooks reduce the full-night sequence length by 64.3\% to 69.7\%, depending on the modality.
These empirical reductions are moderately below the target of 75\%, implied by $N=4$.
This discrepancy arises because chunk boundaries and lengths are adaptively determined based on their context.

\begin{table}[!t]
    \centering
    \caption{Effective compression ratio.}
    \label{tab:compression}
    \resizebox{0.40\linewidth}{!}{%
        \begin{tabular}{cc}
        \toprule
        \textbf{Modality} & \textbf{Compression ratio (\%)}\\
        \midrule
        EEG  & 67.6 \\
        Respiratory  & 64.9 \\
        ECG & 64.3 \\
        PPG & 69.7 \\
        \bottomrule
        \end{tabular}
    }
\end{table}

\clearpage
\section{Generated Spectrogram Samples.}
\begin{figure*}[h]
    \centering
    \begin{subfigure}[t]{0.48\textwidth}
        \centering
        \includegraphics[width=\linewidth,bb=0 0 1146 741]{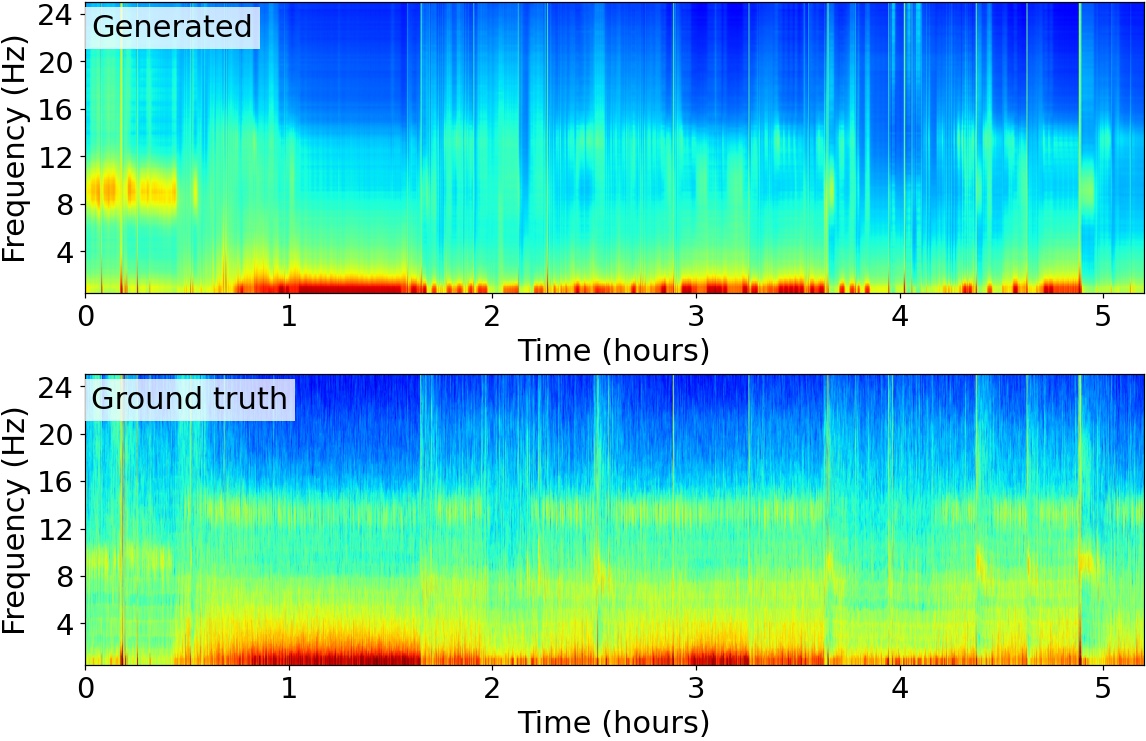}
        \vspace{-2ex}
        \caption{Spectrogram : 64-year-old Female}        
        \label{fig:spec_sample_resp1}
    \end{subfigure}
    \hfill
    \begin{subfigure}[t]{0.48\textwidth}
        \centering
        \includegraphics[width=\linewidth,bb=0 0 1146 741]{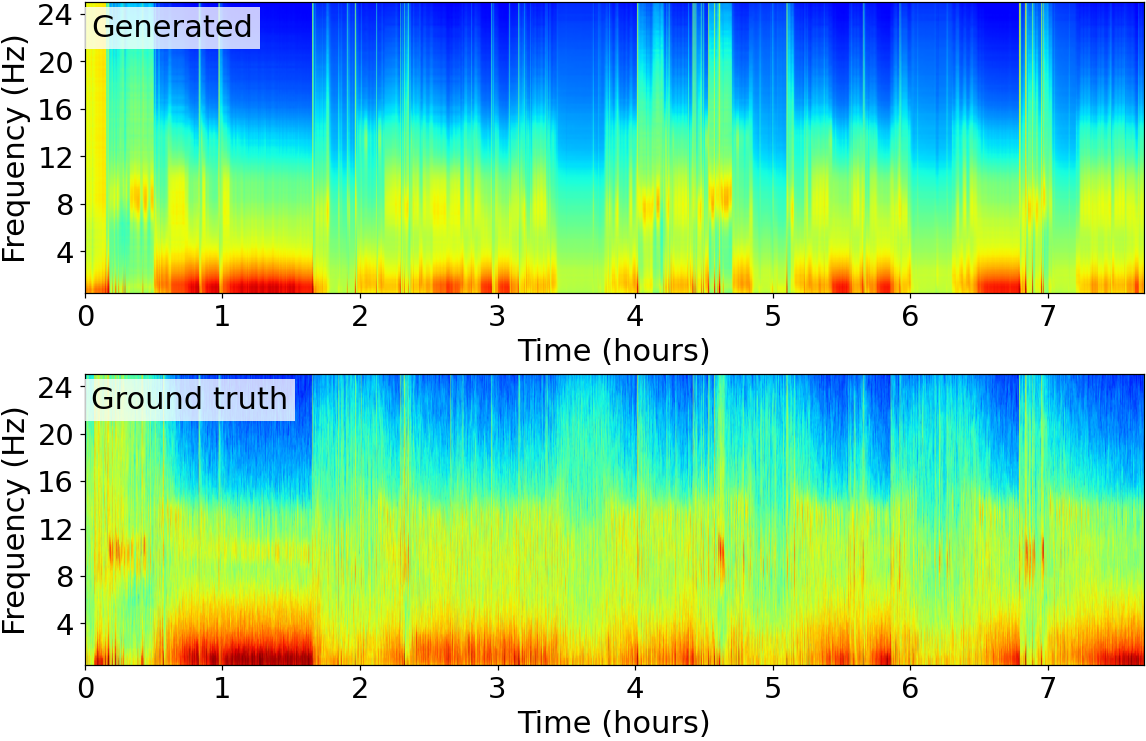}
        \vspace{-2ex}
        \caption{Spectrogram : 65-year-old Female}        
        \label{fig:spec_sample_resp2}
    \end{subfigure}   
    \vspace{-1ex}

    \caption{Generated spectrograms from respiratory.}
    \label{fig:spec_samples_resp}
\end{figure*}

\begin{figure*}[h]
    \centering

    \begin{subfigure}[t]{0.48\textwidth}
        \centering
        \includegraphics[width=\linewidth,bb=0 0 1146 741]{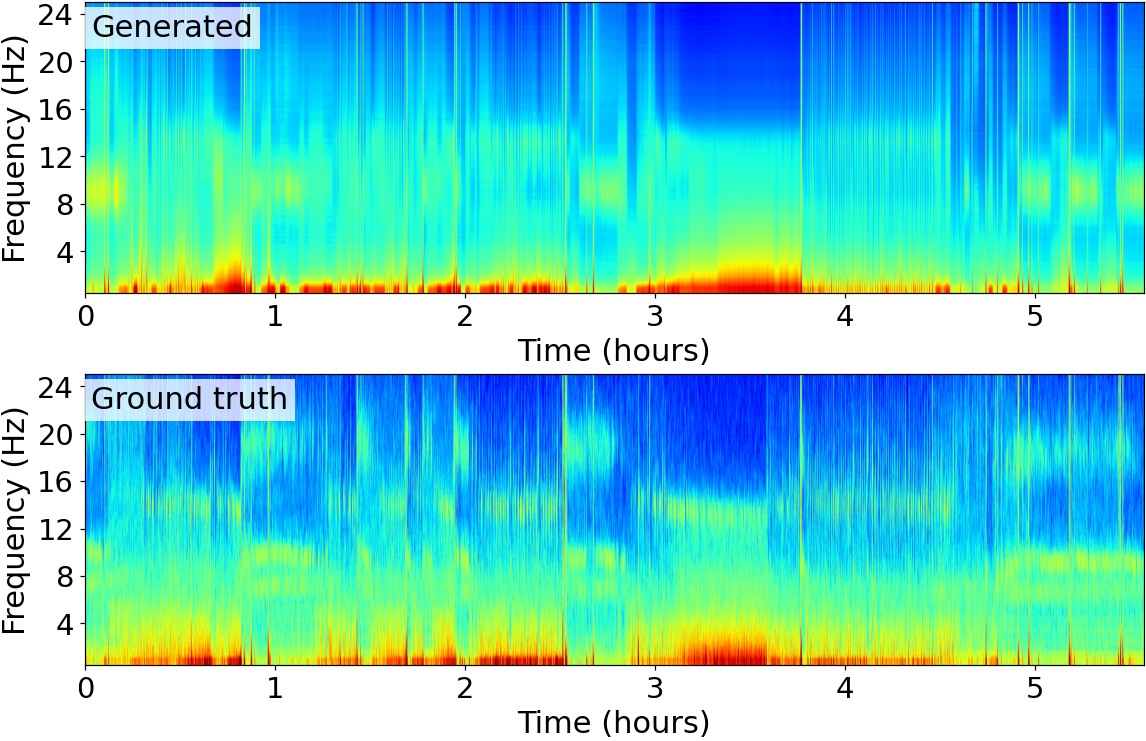}
        \vspace{-2ex}
        \caption{Spectrogram : 46-year-old Male}        
        \label{fig:spec_sample_ecg1}
    \end{subfigure}
    \hfill
    \begin{subfigure}[t]{0.48\textwidth}
        \centering
        \includegraphics[width=\linewidth,bb=0 0 1146 741]{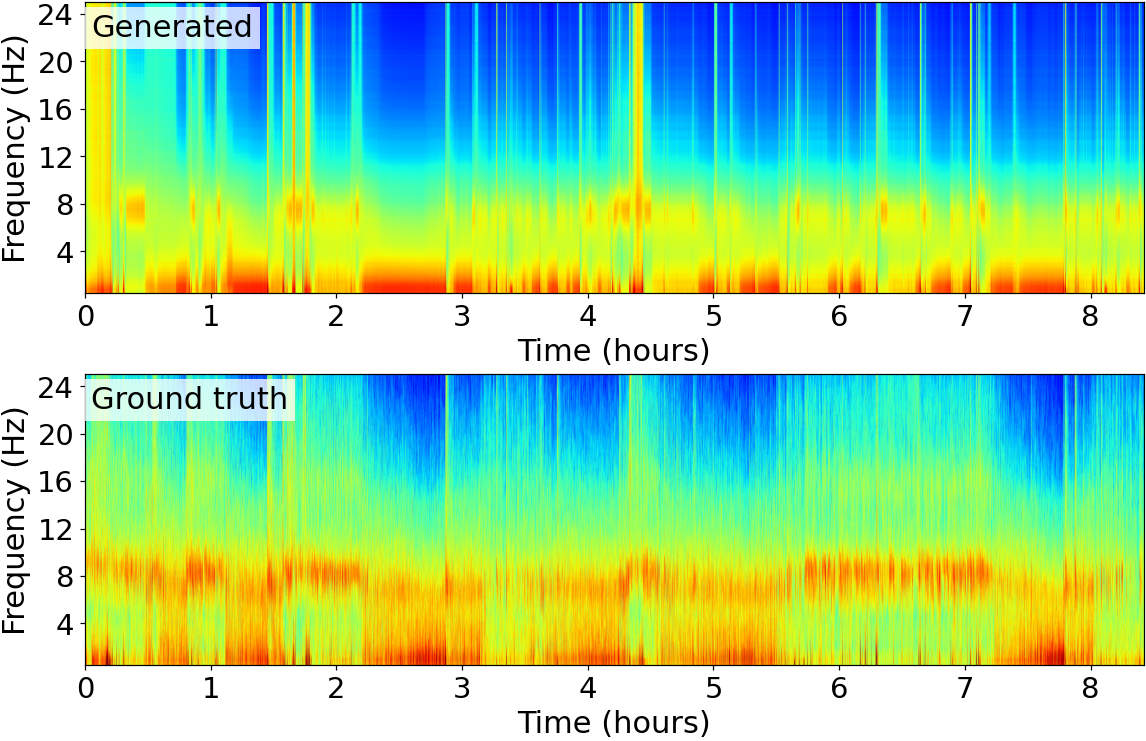}
        \vspace{-2ex}
        \caption{Spectrogram : 79-year-old Female}        
        \label{fig:spec_sample_ecg2}
    \end{subfigure}   
    \vspace{-1ex}
    \caption{Generated spectrograms from ECG.}
    \label{fig:spec_samples_ecg}
\end{figure*}

\begin{figure*}[h]
    \centering

    \begin{subfigure}[t]{0.48\textwidth}
        \centering
        \includegraphics[width=\linewidth,bb=0 0 1146 741]{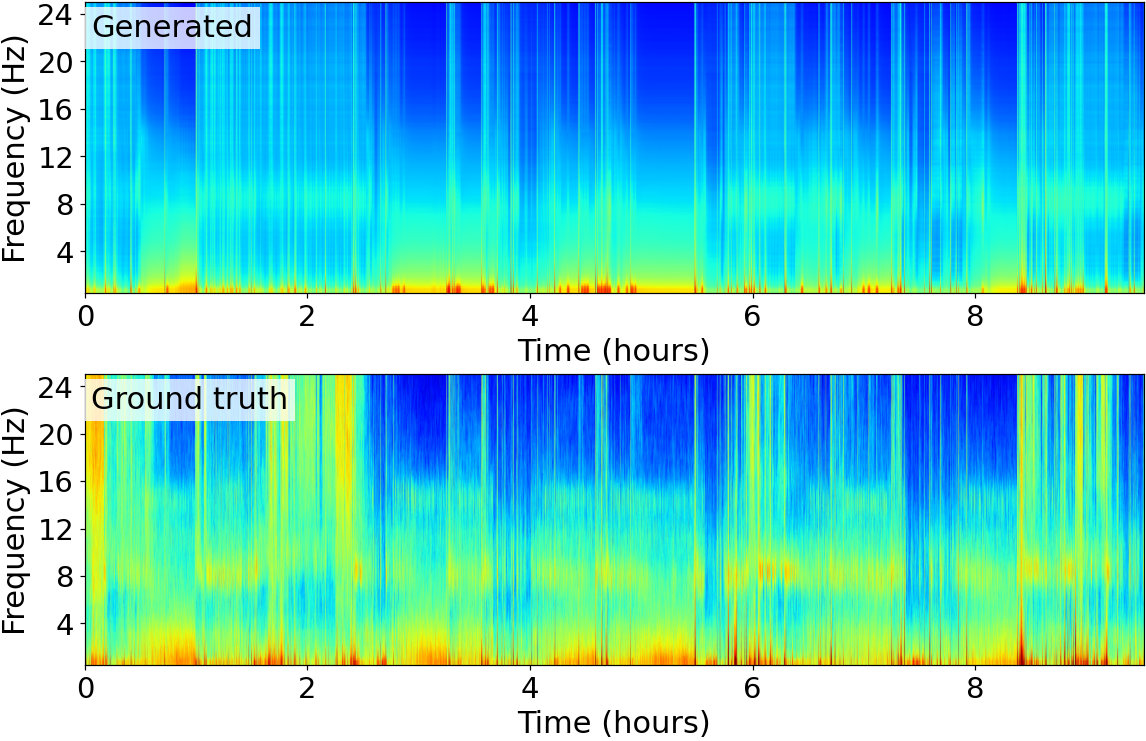}
        \vspace{-2ex}
        \caption{Spectrogram : 55-year-old Male}        
        \label{fig:spec_sample_ppg1}
    \end{subfigure}
    \hfill
    \begin{subfigure}[t]{0.48\textwidth}
        \centering
        \includegraphics[width=\linewidth,bb=0 0 1146 741]{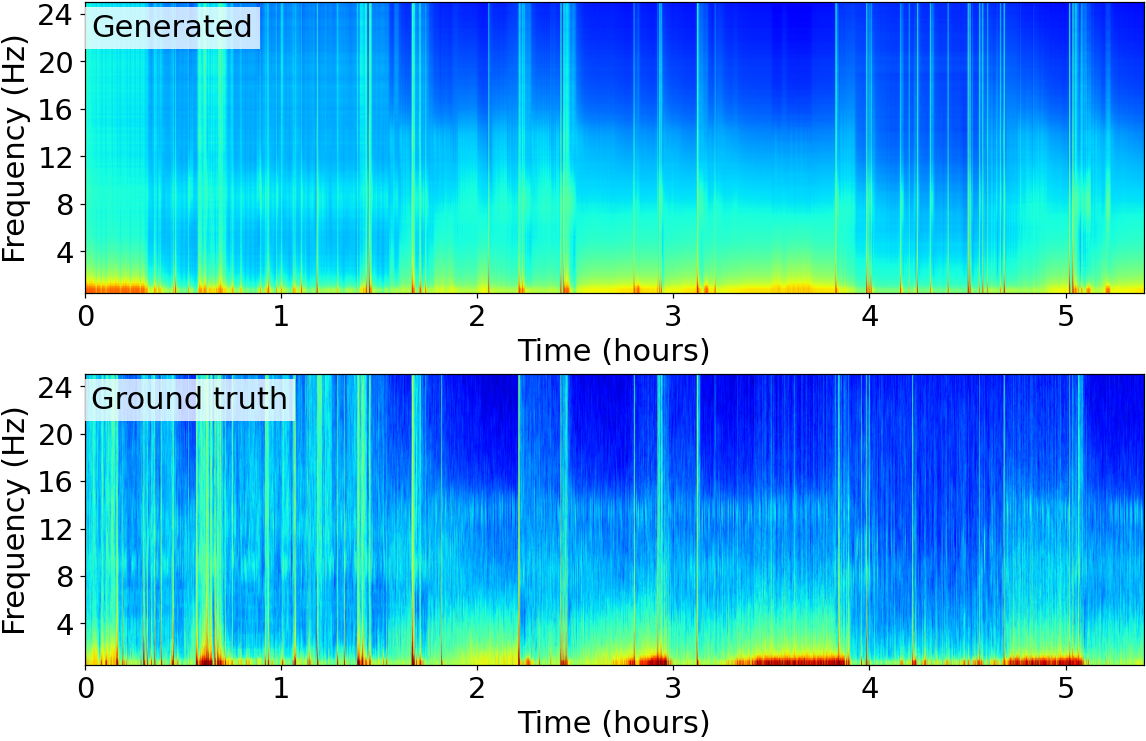}
        \vspace{-2ex}
        \caption{Spectrogram : 65-year-old Male}        
        \label{fig:spec_sample_ppg2}
    \end{subfigure}   
    \vspace{-1ex}
    \caption{Generated spectrograms from PPG.}
    \label{fig:spec_samples_ppg}
\end{figure*}

We provide selected samples of generated spectrograms using respiratory, ECG and PPG, respectively in Figure~\ref{fig:spec_samples_resp}, \ref{fig:spec_samples_ecg}, and \ref{fig:spec_samples_ppg}.

\end{document}